\documentclass[iicol,pdflatex,sn-mathphys-ay]{sn-jnl}

\usepackage{graphicx}%
\usepackage{multirow}%
\usepackage{amsmath,amssymb,amsfonts}%
\usepackage{amsthm}%
\usepackage{mathrsfs}%
\usepackage[title]{appendix}%
\usepackage{xcolor}%
\usepackage{textcomp}%
\usepackage{manyfoot}%
\usepackage{booktabs}%
\usepackage{algorithm}%
\usepackage{algorithmicx}%
\usepackage{algpseudocode}%
\usepackage{listings}%
\usepackage{booktabs}
\usepackage{multirow}
\usepackage{float}
\usepackage{adjustbox}
\usepackage{graphicx}
\usepackage{hyperref}
\usepackage{array}

\theoremstyle{thmstyleone}%
\theoremstyle{thmstyletwo}%

\theoremstyle{thmstylethree}%

\begin{document}

\title[Article Title]{3DCity-LLM: Empowering Multi-modality Large Language Models for 3D City-scale Perception and Understanding}


\author*[1]{\fnm{Yiping} \sur{Chen}}\email{chenyp79@mail.sysu.edu.cn}

\author[1]{\fnm{Jinpeng} \sur{Li}}\email{lijp57@whu.edu.cn}


\author[1]{\fnm{Wenyu} \sur{Ke}}\email{kewy3@mail2.sysu.edu.cn}

\author[1]{\fnm{Yang} \sur{Luo}}\email{luoy583@mail2.sysu.edu.cn}

\author[1]{\fnm{Jie} \sur{Ouyang}}\email{ouyj69@mail2.sysu.edu.cn}

\author[1]{\fnm{Zhongjie} \sur{He}}\email{hezhj53@mail2.sysu.edu.cn}

\author[2]{\fnm{Li} \sur{Liu}}\email{dreamliu2010@gmail.com}

\author[3]{\fnm{Hongchao} \sur{Fan}}\email{hongchao.fan@ntnu.no}

\author[4]{\fnm{Hao} \sur{Wu}}\email{wuhao@ngcc.cn}

\affil*[1]{\orgdiv{School of Geospatial Engineering and Science}, \orgname{Sun Yat-sen University}, \orgaddress{\city{Zhuhai}, \postcode{519082}, \country{China}}}

\affil[2]{\orgdiv{College of Electronic Science}, \orgname{National University of Defense Technology}, \orgaddress{\city{Changsha}, \postcode{410000}, \country{China}}}

\affil[3]{\orgdiv{Department of Civil and Environmental Engineering}, \orgname{Norwegian University of Science and Technology}, \orgaddress{\city{Trondheim}, \country{Norway}}}

\affil[4]{\orgdiv{National Geomatics Center of China}, \orgaddress{\city{Beijing}, \postcode{100830}, \country{China}}}


\abstract{While multi-modality large language models excel in object-centric or indoor scenarios, scaling them to 3D city-scale environments remains a formidable challenge. To bridge this gap, we propose 3DCity-LLM, a unified framework designed for 3D city-scale vision-language perception and understanding. 3DCity-LLM employs a coarse-to-fine feature encoding strategy comprising three parallel branches for target object, inter-object relationship, and global scene. To facilitate large-scale training, we introduce 3DCity-LLM-1.2M dataset that comprises approximately 1.2 million high-quality samples across seven representative task categories, ranging from fine-grained object analysis to multi-faceted scene planning. This strictly quality-controlled dataset integrates explicit 3D numerical information and diverse user-oriented simulations, enriching the question-answering diversity and realism of urban scenarios. Furthermore, we apply a multi-dimensional protocol based on text-similarity metrics and LLM-based semantic assessment to ensure faithful and comprehensive evaluations for all methods. Extensive experiments on two benchmarks demonstrate that 3DCity-LLM significantly outperforms existing state-of-the-art methods, offering a promising and meaningful direction for advancing spatial reasoning and urban intelligence. The source code and dataset are available at \url{https://github.com/SYSU-3DSTAILab/3D-City-LLM}.}




\maketitle

\clearpage
\newpage

\section{Introduction}\label{sec1}

Large language models (LLMs) have rapidly transformed the field of artificial intelligence, demonstrating unprecedented capabilities in reasoning \citep{liu2025logical}, generation \citep{wang2024genartist}, and multi-modality integration \citep{ge2024worldgpt}. By aligning linguistic and visual representations, recent advances such as ChatGPT-5 \citep{achiam2023gpt}, Qwen3 \citep{yang2025qwen3} and LLaVA-Plus \citep{liu2024llava} have shown that language-centric architectures can be adapted for cross-modality understanding. These well-trained models excel at multi-modality tasks, including visual question answering, image captioning and video generation. However, despite their success in small-scale or object-centric scenarios, the potential of LLMs in 3D city-scale environments remains largely unexplored.

Diverse city environments introduce a new level of complexity for multi-modality perception and understanding. Unlike indoor benchmarks that involve a limited number of objects, a city scene usually contains thousands of entities with heterogeneous attributes and intricate spatial relationships. Designing multi-modality large language models (MLLMs) at this scale requires not only recognizing individual objects but also modeling their interactions, functional roles, and contextual significance within the broader urban system. For example, answering a query such as “Which hospital is closest to the railway station? And where is its emergency department located?” requires understanding object categories, precise spatial coordinates, relational proximity, and city scene layout. Such tasks highlight the need for a unified framework that can simultaneously perform 3D object perception, relationship calculation, and holistic scene understanding.

\begin{table*}[t]
\renewcommand{\arraystretch}{1.0}
\centering
\caption{Comparison of representative city-scale vision-language datasets. “N.I.” means the numerical information (3D coordinates, distance, angle, etc.). "C.S." means the contextual simulation for different user groups}
\label{tab:datasets}
\resizebox{1.0\linewidth}{!}{
\begin{tabular}{l l l l l l l}
\toprule
\textbf{Dataset} & \textbf{Source} & \textbf{Modality} & \textbf{N.I.} & \textbf{C.S.} & \textbf{Sample} & \textbf{Task} \\
\midrule
RSVQA \citep{lobry2020rsvqa}  & Sentinel 2, HRO & Text, 2D & $\checkmark$ & $\times$ & 1.1M & QA \\
VQA-TextRS \citep{al2022open} & Google Earth, PatternNet, NWPU & Text, 2D & $\times$ & $\times$ & 6.2k & QA \\
RSGPT \citep{hu2025rsgpt} & DOTA & Text, 2D & $\times$ & $\times$ & 3.5k & QA \\
VRSBench \citep{li2024vrsbench} & DOTA, DIOR & Text, 2D & $\checkmark$ & $\times$ & 205k & Multiple Tasks \\
EarthVQA \citep{wang2024earthvqa} & LoveDA & Text, 2D & $\times$ & $\times$ & 209k & Multiple Tasks \\
UrBench \citep{zhou2025urbench} & Google Street View, Cityscapes, VIGOR & Text, 2D & $\checkmark$ & $\checkmark$ & 11.6k & Multiple Tasks \\
\midrule
CityRefer \citep{miyanishi2023cityrefer} & SensatUrban & Text, 3D & $\times$ & $\times$ & 35.1k & Grounding \\
CityAnchor \citep{licityanchor} & STPLS3D & Text, 2D, 3D & $\checkmark$ & $\times$ & 1.4k & Grounding \\
City-3DQA \citep{sun20243d} & UrbanBIS & Text, 3D & $\times$ & $\times$ & 450k & QA \\
NuScenes-QA \citep{qian2024nuscenes} & NuScenes & Text, 2D, 3D & $\times$ & $\times$ & 460k & QA \\
KITTI360Pose \citep{xia2024text2loc} & KITTI360 & Text, 3D & $\checkmark$ & $\times$ & 43.3k & Localization \\
CityBench \citep{feng2024citybench} & OpenStreetMap, Esri World Imagery & Text, 2D, 3D & $\checkmark$ & $\checkmark$ & 39.0k & Multiple Tasks \\
GeoEval3D \citep{yasuki2025geoprog3d} & GoogleEarth, UrbanScene3D & Text, 2D, 3D & $\checkmark$ & $\checkmark$ & 0.9k & Multiple Tasks \\
UrbanVideo-Bench \citep{zhao2025urbanvideo} & EmbodiedCity, AerialVLN & Text, 2D, 3D & $\checkmark$ & $\checkmark$ & 5.2k & Multiple Tasks \\
Open3DVQA \citep{zhang2025open3dvqa} & EmbodiedCity, UrbanScene3D, WildUAV & Text, 2D, 3D & $\checkmark$ & $\times$ & 73.3k & Multiple Tasks \\
SVM-City \citep{sun2025city} & SensatUrban, UrbanBIS, NuScenes, LoveDA & Text, 2D, 3D & $\times$ & $\times$ & 567k & Multiple Tasks \\
\midrule
\textbf{3DCity-LLM-1.2M (Ours)} & SensatUrban, UrbanBIS, City-BIS & Text, 2D, 3D & $\checkmark$ & $\checkmark$ & \textbf{1.2M} & Multiple Tasks \\
\bottomrule
\end{tabular}}
\end{table*}

As summarized in Table \ref{tab:datasets}, most early benchmarks such as RSVQA \citep{lobry2020rsvqa}, EarthVQA \citep{wang2024earthvqa}, and VRSBench \citep{li2024vrsbench}, are built on 2D aerial or street-view images. Although these datasets are large in scale, they primarily focus on visual question-answering (QA) but lack 3D spatial information required for city scene understanding. Recent resources have moved toward 3D city scenes but suffer from significant constraints: they are either limited to specialized tasks (e.g., grounding or localization) \citep{miyanishi2023cityrefer, xia2024text2loc}, restricted to short-form QA \citep{sun20243d, qian2024nuscenes}, or insufficient for training or fine-tuning LLMs that requires large-scale and high-quality supervision \citep{yasuki2025geoprog3d, zhao2025urbanvideo}.

Beyond the limitations of existing datasets, another pressing challenge lies in evaluation protocol \citep{xu2025pointllm}. As LLMs generate diverse, long-form responses for open-ended urban tasks, traditional text-similarity metrics (e.g., BLEU, ROUGE, METEOR) fail to capture semantic equivalence, especially for complex open-ended questions. As a result, answers that are logically coherent and factually accurate may be penalized simply because they adopt different wording, sentence structure, or narrative style from the ground truth. For example, one answer may describe spatial relations in quantitative terms (“45 meters southwest”), while another may rely on landmarks (“next to the parking lot on the southwest side”). Both of the answers are correct, yet traditional metrics would assign them with different scores. Therefore, relying solely on such text-similarity evaluations risks undervaluing legitimate diversity, obscuring a model’s reasoning ability, and ultimately providing an incomprehensive assessment of its real-world applicability.

In this paper, we introduce 3DCity-LLM, a unified framework designed to extend LLMs to 3D city-scale vision-language perception and understanding. Unlike existing models that focus on indoor or object-centric scenarios, 3DCity-LLM addresses the challenges of massive urban objects and intricate spatial dependencies through a novel coarse-to-fine feature encoding strategy. This mechanism integrates object-level geometry, inter-object relationship topology, and global scene semantics into a shared embedding space. Through task-driven instruction tuning, 3DCity-LLM is able to handle diverse tasks ranging from fine-grained object analysis to complex scene analysis and goal-oriented planning. To support this framework, we construct the 3DCity-LLM-1.2M dataset, comprising approximately 1.2 million high-quality samples across a systematic taxonomy of seven task categories. This dataset is generated through an automated pipeline that extracts structured city scene attributes including explicit 3D numerical information and employs advanced vision-language models (VLMs) for instruction-driven generation enhanced by contextual simulations of various user personas. Then, we implement a rigorous quality control process to eliminate hallucinations, remove potential privacy-sensitive information, and ensure linguistic clarity. Furthermore, we address the limitations of traditional text-similarity metrics in evaluating long-form textual answers by introducing a multi-dimensional evaluation protocol. This protocol complements standard metrics like BLEU and ROUGE with LLM-based comprehensive assessments of logicality and reliability, explicitly measuring internal coherence and factual alignment with 3D scene evidence to ensure a faithful assessment of model performance in open-ended city-scale tasks.

Our contributions can be summarized as follows:

(1) We propose 3DCity-LLM, a unified framework designed for 3D perception and understanding within city environments. This framework leverages a coarse-to-fine feature encoding strategy and a task-driven instruction tuning paradigm to jointly support object-level, relationship-level, and scene-level tasks.

(2) We introduce the 3DCity-LLM-1.2M dataset, a large-scale dataset containing approximately 1.2 million samples with explicit 3D numerical information and diverse user-oriented contextual simulation. The dataset establishes a systematic taxonomy of seven representative tasks, including object caption, object localization, object analysis, relationship computation, scene caption, scene analysis and scene planning.

(3) We conduct sufficient experiments with comprehensive evaluation protocol that combines text-similarity metrics with LLM-based semantic assessments, evaluating the performances of proposed 3DCity-LLM and other competitive methods. The experiments demonstrate that 3DCity-LLM achieves consistent performance improvements, with gains ranging from 0.50 to 8.40 in BLEU-4, 1.07 to 10.69 in METEOR, and 0.16 to 1.51 in reliability.

\section{Related Work}\label{sec2}

\subsection{3D Vision-Language Dataset}

The rapid advancement of 3D vision-language (3D-VL) research has intensified the demand for datasets that link spatial understanding with natural language, resulting in a new generation of well annotated 3D-language datasets. Existing efforts can be broadly classified into two domains: indoor datasets that focus on confined scenes, and outdoor datasets designed for large and complex environments.

\textbf{Indoor Dataset.} Early research in this domain has developed task-specific benchmarks for 3D captioning \citep{chen2021scan2cap}, question answering \citep{azuma2022scanqa, ma2022sqa3d}, and visual grounding \citep{chen2020scanrefer}. While these datasets have advanced individual tasks, they are typically limited in scope and lack cross-task generalization, restricting 3D LLMs from transferring knowledge effectively across different 3D-VL tasks. To address this limitation, recent works \citep{wang2024embodiedscan, yang20253d} have introduced large-scale, multi-purpose datasets that support a broader spectrum of 3D-VL applications, encompassing dense captioning, multi-turn dialogue, robotic navigation, and embodied planning.

\textbf{Outdoor Dataset.} 3D-VL datasets in outdoor environments remain scarce and fragmented. The existing resources primarily focus on a limited set of subtasks. CityRefer \citep{miyanishi2023cityrefer} provides a benchmark for visual grounding in urban point clouds, City-3DQA \citep{sun20243d} emphasizes large-scale QA for city scenes but limits responses to short tokens, hindering long-form reasoning. NuScenes-QA \citep{qian2024nuscenes} adapts autonomous-driving scenes for street-view QA but remains centered on instance-level perception. Recent works such as CityBench \citep{feng2024citybench}, GeoEval3D \citep{yasuki2025geoprog3d}, UrbanVideo-Bench \citep{zhao2025urbanvideo} and Open3DVQA \citep{zhang2025open3dvqa} attempt to expand task diversity, yet their annotation size remain limited. The latest effort SVM-City \citep{sun2025city} advances multi-task learning via multi-source data integration. However, its dependency on static generation templates (e.g., localization, measurement, functionality, and logical reasoning) results in syntactic homogeneity, which restricts linguistic diversity and the capacity for open-ended urban understanding.

\begin{table*}[ht]
\renewcommand{\arraystretch}{1.0}
\centering
\caption{Statistical overview of the existing 3D large-scale vision-language datasets designed for city scenes. We report the number of samples, average question length (\textit{Que. Len.}) and answer length (\textit{Ans. Len.}) across different datasets. 'Roadside’ denotes vehicle-mounted perspective data, 'Urban-scale' refers to city-level data reconstructed from aerial LiDAR or photogrammetry. The SVM-City dataset is currently unavailable for public access}
\resizebox{0.9\linewidth}{!}{
\begin{tabular}{l|c|c|cc|cc}
\toprule
\multirow{2}{*}{\textbf{Dataset}} & \multirow{2}{*}{\textbf{Area}} & \multirow{2}{*}{\textbf{Sample}} & \multicolumn{2}{c|}{\textit{Word}} & \multicolumn{2}{c}{\textit{Token}} \\
 & & & \textit{Que. Len.} & \textit{Ans. Len.} & \textit{Que. Len.} & \textit{Ans. Len.} \\
\midrule
NuScenes-QA \citep{qian2024nuscenes} & Roadside & 460k & 14.32 & 1.04 & 15.59 & 1.05 \\
City-3DQA \citep{sun20243d} & Urban-scale & 450k & 12.56 & 1.80 & 14.80 & 1.81 \\
SVM-City \citep{sun2025city} & Urban-scale & 567k & - & - & - & - \\
3DCity-LLM-1.2M (ours) & Urban-scale & \textbf{1.2M} & 13.49 & \textbf{39.47} & 17.29 & \textbf{49.44} \\
\bottomrule
\end{tabular}
}
\label{table:dataset}
\end{table*}

\begin{table*}[ht]
\centering
\caption{Statistical overview of 3DCity-LLM-1.2M dataset. We report the number of samples, average question length (\textit{Que. Len.}) and answer length (\textit{Ans. Len.}) across different task categories}
\resizebox{0.90\linewidth}{!}{
\begin{tabular}{l|c|cc|cc}
\toprule
\multirow{2}{*}{\textbf{Task Category}} & \multirow{2}{*}{\textbf{Sample (Proportion)}} & \multicolumn{2}{c|}{\textit{Word}} & \multicolumn{2}{c}{\textit{Token}} \\
 & & \textit{Que. Len.} & \textit{Ans. Len.} & \textit{Que. Len.} & \textit{Ans. Len.} \\
\midrule
Object Caption           & 350k (28.3\%) & 10.32 & 60.46 & 12.79 & 73.89 \\
Object Localization      & 94k (7.6\%) & 16.48 & 50.64 & 18.34 & 73.35 \\
Object Analysis          & 470k (37.9\%) & 11.25 & 23.42 & 14.07 & 28.63 \\
Relationship Computation & 56k (4.6\%)  & 25.45 & 40.62 & 53.83 & 65.21 \\
Scene Caption            & 160k (12.9\%) & 9.73 & 32.40 & 11.33 & 38.56 \\
Scene Analysis           & 55k (4.5\%)   & 17.86 & 31.24 & 19.50 & 35.91 \\
Scene Planning           & 52k (4.2\%)   & 41.14 & 52.53 & 47.36 & 61.10 \\
\bottomrule
\end{tabular}
}
\label{table:city_llm_stats}
\end{table*}

\subsection{Multi-modality LLMs}

LLMs such as ChatGPT \citep{achiam2023gpt} and LLaMA \citep{touvron2023llama} have exhibited remarkable generalization ability across various language-based tasks by leveraging self-supervised pre-training paradigms. Motivated by the exceptional versatile capabilities, researchers are investigating methodologies to extend LLMs into understanding 2D images \citep{awais2025foundation} and 3D spatial contexts \citep{ma2024llms}. 

\textbf{2D LLMs.} The recent development of LLMs \citep{achiam2023gpt, touvron2023llama} has spurred rapid advances in 2D LLMs \citep{lai2024lisa, liu2023visual}, which extend vision-language comprehension by integrating a visual encoder and a cross-modal projector. Through visual instruction tuning, 2D LLMs establish robust correspondences between text descriptions and visual representations, enabling a broad range of downstream tasks such as image captioning, visual question answering, and open-vocabulary segmentation. On the basis of text-image alignment, recent studies \citep{dong2025insight, thawakar2025llamav} have incorporated multi-turn dialogue and multi-step chain-of-thought (CoT) to enhance context-dependent reasoning grounded in visual information, thereby achieving more nuanced understanding of object attributes and topological dependencies.

\textbf{3D LLMs.} Integrating 3D content into LLMs has become a fundamental step toward achieving natural and interactive language understanding within spatially grounded real-world environments. Early 3D point cloud LLMs \citep{xu2024pointllm, wang2023chat} directly align geometric point representations with textual queries. While these approaches demonstrate strong performance on object-level perception, they struggle to capture holistic scene semantics and contextual dependencies. 3D Vision-Language Models (3D-VLMs) such as LLaVA-3D \citep{zhu2024llava} leverage pretrained 2D foundation models (e.g., LLaVA \citep{liu2023visual}) as backbones to extract richly visual features from multi-view images, which are fused with 3D geometry-aware features derived from point clouds to yield unified cross-modal representations capable of describing spatially coherent scenes. Another emerging trend involves the adoption of video-based LLMs \citep{qi2025gpt4scene, zheng2025video} for 3D-VL tasks, leveraging rich textures and temporal continuity in videos to infer implicit 3D geometry from motion, providing a more complete spatial understanding for indoor scenes. However, scaling 3D LLMs to city-level environments presents fundamental challenges beyond object-centric or room-scale domains, as it entails comprehensive 3D city-scale spatial understanding that integrates object attributes, relational topology, and scene layout.

\begin{figure*}[]
\centering
\vspace{-0.4cm}
\includegraphics[width=0.99\textwidth]{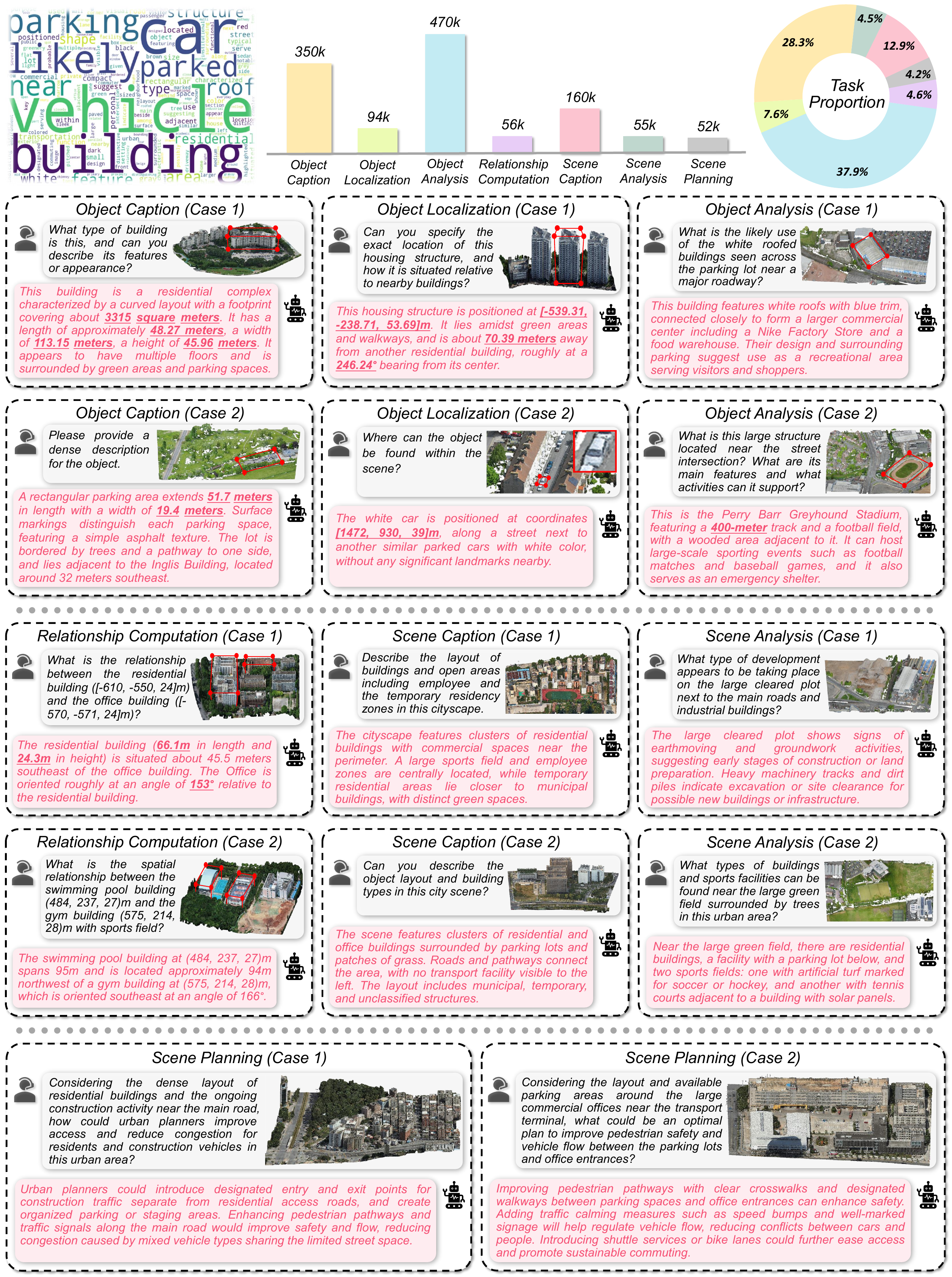}
\vspace{-0.0cm}
\caption{The statistical information and representative cases of 3DCity-LLM-1.2M dataset. We provide two representative cases for each task type. The specific 3D numerical information is highlighted with underscores}
\label{Fig2-DataExample}
\end{figure*}

\section{Dataset Construction}\label{sec3}

In this section, we introduce the self-constructed 3DCity-LLM-1.2M dataset, containing 1.2 million samples designed to advance city-scale perception and understanding in urban environments. The representative examples of 3DCity-LLM-1.2M dataset are shown in Fig. \ref{Fig2-DataExample}. The detailed statistical information is presented in Table \ref{table:dataset} and Table \ref{table:city_llm_stats}. We define the task taxonomy to highlight the scope and characteristics of the dataset, followed by an overview of the automated data generation pipeline and quality control process.

\subsection{Task Definition}

We propose a hierarchical taxonomy of 3D vision–language tasks that spans fine-grained object perception, relational reasoning, and holistic scene interpretation. This taxonomy includes seven representative categories:

$\bullet$ \textit{Object Caption (350k): Generate concise or detailed textual descriptions for individual urban objects, capturing their visual appearance, semantic category, and unique attributes.}

$\bullet$ \textit{Object Localization (94k): Retrieve the spatial coordinates or segmentation mask of target urban object based on textual descriptions.}

$\bullet$ \textit{Object Analysis (470k): Provide in-depth interpretations of urban objects, including their physical characteristics, functional roles, and potential affordances within the urban environment.}

$\bullet$ \textit{Relationship Computation (56k): Quantitatively infer spatial relationships and topological dependencies between two or more urban objects.}

$\bullet$ \textit{Scene Caption (160k): Generate concise or detailed textual descriptions that summarize entire city scenes, capturing object compositions, spatial arrangements, and contextual cues.}

$\bullet$ \textit{Scene Analysis (55k): Perform high-level interpretation by integrating object-centric, relational, and contextual information to support comprehensive understanding within complex city environments.}

$\bullet$ \textit{Scene Planning (52k): Conduct goal-oriented reasoning and decision-making based on scene understanding, involving formulating feasible plans, evaluating accessibility, optimizing routes, and proposing spatial interventions.}

This hierarchical taxonomy provides a comprehensive coverage of 3D perception and understanding tasks. Importantly, most existing city-scale benchmarks can be categorized into one or more task types within this taxonomy, underscoring its extensibility. 

\begin{figure*}[]
\centering
\includegraphics[width=1.0\textwidth]{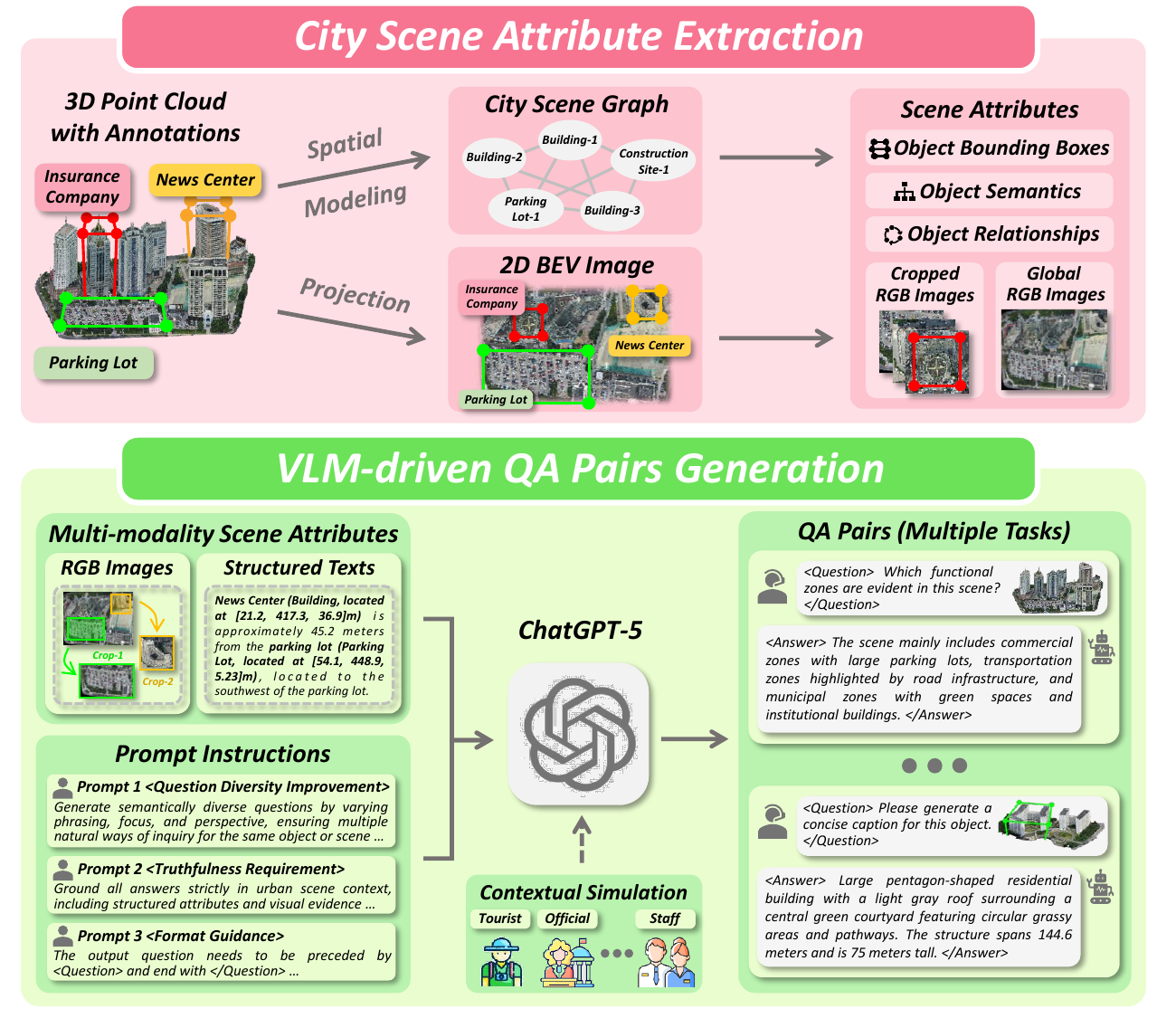}
\vspace{-0.5cm}
\caption{Automated data generation pipeline of 3DCity-LLM-1.2M dataset, leveraging instance-level masks and landmark annotations from the SensatUrban, UrbanBIS, and City-BIS datasets. We first construct city scene attributes for each environments as input for VLM, then prompt the VLM to generate diverse and high-quality QA pairs grounded in the scene with multiple instructions. In total, we built 1.2M samples spanning object caption, object localization, object analysis, relationship computation, scene caption, scene analysis and scene planning tasks}
\label{Fig3-DataGeneration}
\end{figure*}

\subsection{Automated Data Generation}

As shown in Fig. \ref{Fig3-DataGeneration}, to construct the 3DCity-LLM-1.2M dataset, we develop an automated pipeline that includes two stages: (1) City scene attribute extraction for establishing adequate and reliable scene context. (2) VLM-driven QA generation based on the multi-modality scene attributes and well-designed prompt instructions.

\subsubsection{City Scene Attribute Extraction}

We employ city-scale point clouds from the SensatUrban \citep{hu2022sensaturban}, UrbanBIS \citep{yang2023urbanbis}, and City-BIS \citep{li2026city} datasets, exploiting their integrated instance-level masks and landmark annotations as input. Object instances are organized into a city scene graph, where nodes represent objects and edges encode relationships such as adjacency, containment, and orientation. In parallel, the point cloud is projected into a 2D Bird’s-Eye View (BEV) image, with each object highlighted by a unique identifier.

Given that VLM can only directly process text and image modalities, we convert scene attributes into structured texts and RGB images. On the visual side, we crop object-centric RGB images and pair them with global scene views. On the textual side, we serialize structured attributes into descriptive templates that simultaneously capture object landmarks, spatial coordinates, and pairwise distances (e.g., \textit{“News Center (Building, located at [21.2, 417.3, 36.9]m) is approximately 45.2 meters from the parking lot (Parking Lot, located at [54.1, 448.9, 5.23]m), located to the southwest of the parking lot.”}). These structured texts and RGB images form multi-modality scene attributes that can be directly ingested by the VLM for QA pairs generation.


\subsubsection{VLM-driven QA Pair Generation}

With scene attributes prepared, the advanced VLMs (e.g., ChatGPT-5 \citep{achiam2023gpt}) are employed to generate QA pairs for multiple tasks. 

$\bullet$ \textbf{Diversity Improvement.} To encourage lexical and syntactic diversity, each generated QA pair undergoes question diversification. The VLM is instructed to reformulate the same query into multiple phrasings while preserving semantic intent. For example, \textit{“Where is the nearest hospital?”} may also appear as \textit{“Which location in the city corresponds to the closest hospital?”} or \textit{“Can you identify the hospital closest to this current place?”}. 

$\bullet$ \textbf{Truthfulness Requirement.} Answers must remain grounded in the provided city scene attributes. The VLM receives structured texts containing object landmarks, attributes, and precise spatial positions, along with RGB images views as scene context references. The generated answers must explicitly reference these elements, avoiding introducing unsupported assumptions or potential hallucinations.

$\bullet$ \textbf{Format Guidance.} Each output question and answer strictly includes pre-defined identifier marks. The question should begin with $<Question>$ and end with $</Question>$, while the answer should begin with $<Answer>$ and end with $</Answer>$.

$\bullet$ \textbf{Contextual Simulation.} To enhance the contextual relevance and linguistic realism of the generated QA pairs, the VLM is prompted to assume various personas such as a tourist, a government official, or a company staff. Each persona is associated with a distinct linguistic style, reasoning depth, and attention focus. For instance, a tourist may generate casual questions regarding scenic spots or landmarks (e.g., ``What famous attractions can I visit near this area?''), whereas a government official may pose formal analytical queries concerning environmental impact (e.g., ``How does this district’s building layout comply with current development regulations?'').

To guide this role-conditioned prompting, we employ a template-based learning strategy. Several examples are first constructed for each persona, illustrating the expected tone, vocabulary, and reasoning pattern. These examples are then added in the instruction prompt as few-shot demonstrations, enabling the VLM to adapt its phrasing and inference process according to the simulated context.

\subsection{Quality Control}


To ensure the quality of 3DCity-LLM-1.2M dataset, we adopt an automated cross-checking procedure on the validation set by leveraging multiple VLMs. Specifically, ChatGPT-5 \citep{achiam2023gpt}, Gemini 2.5 \citep{comanici2025gemini} and Claude-3.5-Sonnet\footnote{\url{https://www.anthropic.com/news/3-5-models-and-computer-use}} are employed as independent evaluators. Each model is prompted to identify potential deficiencies in validation samples, such as: 

$\bullet$ \textit{Residual template artifacts in questions or answers.}

$\bullet$ \textit{Risk of compromising personal privacy in questions or answers.}

$\bullet$ \textit{Ambiguous or ill-posed question formulations.}

$\bullet$ \textit{Uninformative, illogical or overly short answers.}

$\bullet$ \textit{Inconsistencies between answers and provided scene attributes.}




\section{Method}\label{sec4}

\begin{figure*}[]
\centering
\includegraphics[width=1.0\textwidth]{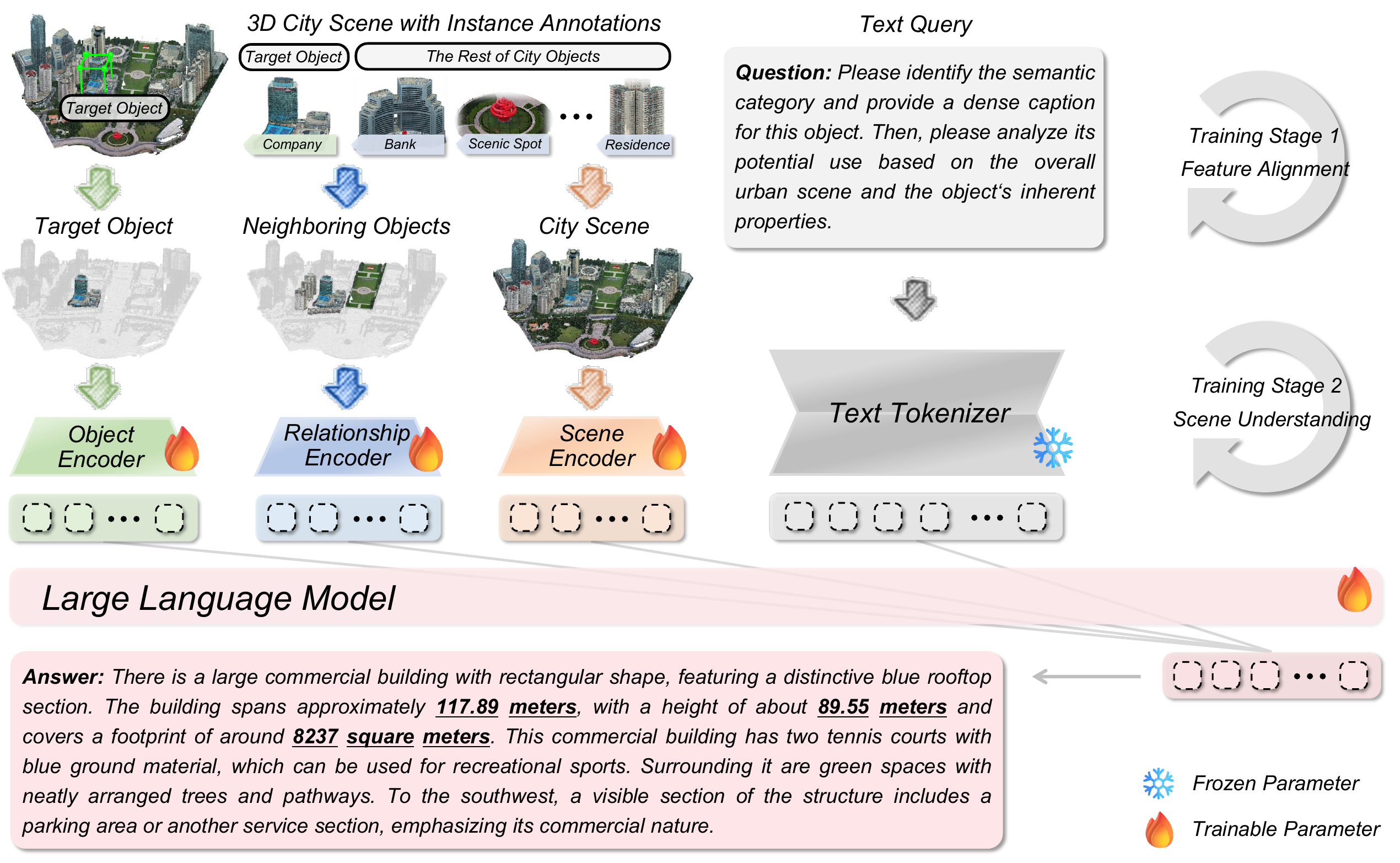}
\vspace{-0.4cm}
\caption{Model architecture of 3DCity-LLM. 3DCity-LLM receives target object, its neighboring objects, city scene and text query as multi-modality inputs, then identifies task type based on the text query and activates the corresponding feature encoding branches before producing the final answers}
\label{Fig4-CityLLM-Method}
\end{figure*}

\subsection{3DCity-LLM Framework}

As shown in Fig. \ref{Fig4-CityLLM-Method}, 3DCity-LLM is built upon a powerful LLM backbone and extended to handle 3D-VL understanding in city-scale environments. To achieve this, 3DCity-LLM introduces a coarse-to-fine feature encoding mechanism that jointly models the target object, its neighboring objects, and city scene. These hierarchical representations are fused into the LLM through feature projectors as modality-aware embeddings. Furthermore, 3DCity-LLM follows an instruction-driven paradigm, where task-specific objectives are formulated through well-designed instruction parser, allowing 3DCity-LLM to flexibly switch between diverse 3D-VL tasks without architectural modification.

\subsubsection{Coarse-to-fine Feature Encoding}

To achieve hierarchical 3D understanding in city-scale environments, 3DCity-LLM adopts a coarse-to-fine feature encoding that extracts multi-modality cues. Formally, given text query $T$ and city scene $\mathcal{S}$ containing $\mathcal{N}$ objects $\{o_i\}_{i=1}^{\mathcal{N}}$ segmented by 3D instance segmentation model SoftGroup \citep{vu2022softgroup}, 3DCity-LLM perform text, object, relationship and scene encoding simultaneously through four branches:

\textbf{Text Encoding.} We tokenize the language instruction $T$ to text feature vector ${E}_{T} \in \mathbb{R}^{l \times d}$, where $l$ is a pre-defined sentence length and $d = 1024$ is the feature dimension.

\textbf{Object Encoding.} For target object $o_t$, 3DCity-LLM extract three object-centric feature streams: 

\begin{itemize}
    \item \textit{Local View Feature.} We project $o_t$ into a top-view RGB patch and extract object-centric visual features $\mathbf{f}_{\text{v}} \in \mathbb{R}^{C\times d}$ using a pretrained CLIP encoder \citep{radford2021learning}.  
    \item \textit{Shape Feature.} We sample 3D points from the target object and process them through Uni3D \citep{zhou2023uni3d} to obtain a shape-aware feature $\mathbf{f}_{\text{s}} \in \mathbb{R}^{1\times d}$ with spatial structure information.  
    \item \textit{Landmark Feature.} If an object in city scene has its landmark name (e.g., ``City Hall'' or ``St. John’s Tower''), we embed the name into $\mathbf{f}_{\text{l}} \in \mathbb{R}^{1\times d}$ via BERT \citep{devlin2019bert} model and and set $\mathbf{f}_{\text{l}}$ to zeros otherwise.
\end{itemize}

After feature concatenation, we apply an object feature projector $\mathrm{Proj}_o(\cdot)$ to transform the fused feature $[\mathbf{f}_{\text{v}};\mathbf{f}_{\text{s}};\mathbf{f}_{\text{l}}]$ into the LLM’s embedding space:

\begin{equation}
\mathbf{E}_{o} = \mathrm{Proj}_o([\mathbf{f}_{\text{v}};\mathbf{f}_{\text{s}};\mathbf{f}_{\text{l}}]).
\end{equation}

This projection yields the object embeddings $\mathbf{E}_{o} \in \mathbb{R}^{(C+2)\times4096}$ that are aligned with language feature manifold.

\begin{figure*}[]
\centering
\includegraphics[width=1.0\textwidth]{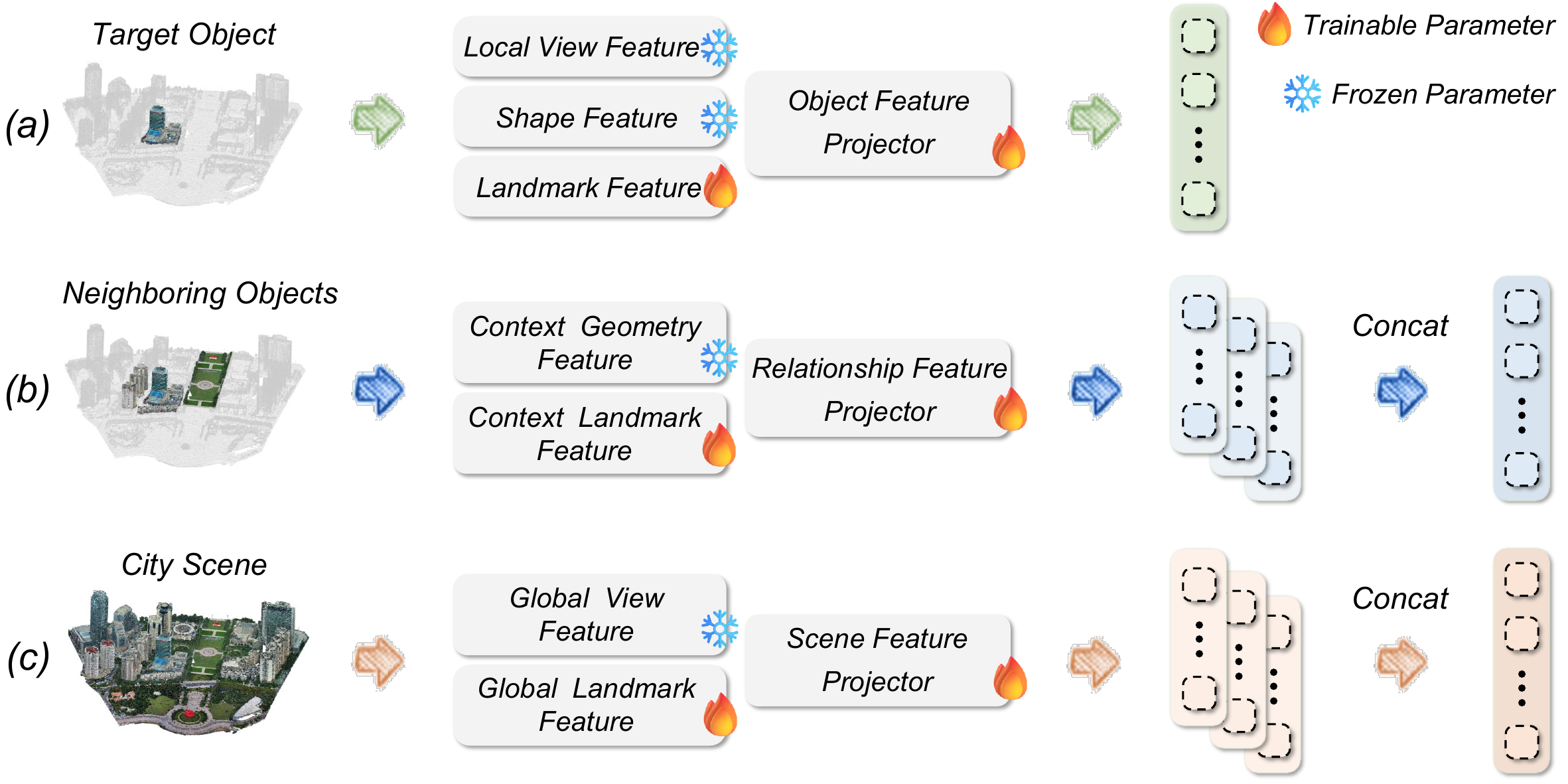}
\vspace{-0.4cm}
\caption{The coarse-to-fine feature encoding in 3DCity-LLM. (a) Object Encoding (b) Relationship Encoding (c) Scene Encoding}
\label{Fig5-CityLLM-Encoding}
\end{figure*}

\textbf{Relationship Encoding.} Descriptions of a target object often rely on its spatial relations with surrounding entities. To capture such relationship context, we perform K-nearest neighbor (KNN) search around the target object $o_t$ to retrieve its neighboring set $\mathcal{N}_t = \{o_k\}_{k=1}^{K}$ based on 3D Euclidean distance. 3DCity-LLM extracts two context feature streams:

\begin{itemize}
    \item \textit{Context Geometry Feature.} 
    Each neighboring object $o_k$ is encoded by Uni3D \citep{zhou2023uni3d} into a shape-aware feature $\mathbf{f}^{(k)}_{\text{s}} \in \mathbb{R}^{1\times d}$, forming the feature set $\mathbf{F}_{\text{s}} = \{\mathbf{f}^{(k)}_{\text{s}}\}_{k=1}^{K} \in \mathbb{R}^{K\times d}$. To enhance spatial awareness, the relative position between each neighbor and the target object is computed as $\Delta \mathbf{p}^{(k)} = \mathbf{p}_{o_k} - \mathbf{p}_{o_t}$ and encoded via learnable function. The attention weight $\{\alpha_k\}_{k=1}^{K}$ between the target object and its $K$ neighbors is defined as:
    \begin{equation}
    \alpha_k = 
    \frac{\exp\!\big(\mathbf{f}^{(t)}_{\text{s}} \cdot [\mathbf{f}^{(k)}_{\text{s}} + \phi(\Delta \mathbf{p}^{(k)})]\big)}
    {\sum_{j=1}^{K}\exp\!\big(\mathbf{f}^{(t)}_{\text{s}} \cdot [\mathbf{f}^{(j)}_{\text{s}} + \phi(\Delta \mathbf{p}^{(j)})]\big)}.
    \end{equation}

    The $\alpha_k$ indicate the geometric relevance of each neighboring object $o_k$ to the target object. The context geometric feature $\mathbf{F}_{\text{g}} \in \mathbb{R}^{K\times d}$ is obtained as:
    \begin{equation}
    \mathbf{F}_{\text{g}} = \{\alpha_k \, \mathbf{f}^{(k)}_{\text{s}}\}_{k=1}^{K}.
    \end{equation}

    \item \textit{Context Landmark Feature.}
    For the $K$ neighboring objects, if a landmark name exists, it is encoded using BERT \citep{devlin2019bert} model to obtain $\mathbf{f}^{(k)}_{\text{l}} \in \mathbb{R}^{1\times d}$. Then, all landmark features from the neighboring object set are stacked to form the context landmark feature $\mathbf{F}_{\text{l}} \in \mathbb{R}^{K\times d}$. 

\end{itemize}

After obtaining the context features, we concatenate them along the token dimension to form a fused representation $[\mathbf{F}_{\text{g}}; \mathbf{F}_{\text{l}}] \in \mathbb{R}^{(2K)\times d}$, which is passed through a relationship feature projector $\mathrm{Proj}_r(\cdot)$:

\begin{equation}
\mathbf{E}_{r} = \mathrm{Proj}_r([\mathbf{F}_{\text{g}}; \mathbf{F}_{\text{l}}]).
\end{equation}

This projection yielding the relationship embeddings $\mathbf{E}_{r} \in \mathbb{R}^{(2K)\times4096}$ that are aligned with language feature manifold.

\textbf{Scene Encoding.} For city scene $\mathcal{S}$, 3DCity-LLM extracts two global feature streams:

\begin{itemize}
    \item \textit{Global View Feature.} We project $\mathcal{S}$ into a global top-view RGB map and extract its visual representation using a pre-trained CLIP encoder \citep{radford2021learning}, yielding a global visual feature $\mathbf{F}_{\text{v}}^{\text{Sce}} \in \mathbb{R}^{C\times d}$.

    \item \textit{Global Landmark Feature.} We collect all objects with valid landmark names and encode each name with BERT \citep{devlin2019bert} model to obtain global landmark feature set $\mathbf{F}_{\text{l}}^{\text{Sce}} = \{\mathbf{f}_{\text{l}}^{(m)}\}_{m=1}^{M} \in \mathbb{R}^{M\times d}$, where $M$ is the number of objects with landmark names in the scene.
\end{itemize}

Both the global view feature and landmark feature set are concatenated along the token dimension and projected into the LLM embedding space through a scene feature projector $\mathrm{Proj}_s(\cdot)$:

\begin{equation}
\mathbf{E}_{s} = \mathrm{Proj}_s([\mathbf{F}_{\text{v}}^{\text{Sce}}; \mathbf{F}_{\text{l}}^{\text{Sce}}]),
\end{equation}

resulting in the scene embeddings $\mathbf{E}_{s} \in \mathbb{R}^{(C+M)\times4096}$.

\subsubsection{Task Instruction}

3DCity-LLM adopts an instruction-driven paradigm that determines the task type and the corresponding activated feature branches. Each input consists of text query $T$ and city scene $\mathcal{S}$ containing $\mathcal{N}$ objects $\{o_i\}_{i=1}^{\mathcal{N}}$. The instruction parser identifies whether the input corresponds to an object-level, relationship-level, or scene-level task and selects the relevant target set $\mathcal{O}_t \subseteq \mathcal{S}$. 

For object-level and relationship-level tasks, one or more target objects $o_t$ are manually selected. 3DCity-LLM encodes the corresponding object embedding $\mathbf{E}_{o}$, relationship embedding $\mathbf{E}_{r}$, and the scene embedding $\mathbf{E}_{s}$. For scene-level tasks, no object selection is required. 3DCity-LLM directly encodes the scene embedding $\mathbf{E}_{s}$, while setting the object and relationship embeddings ($\mathbf{E}_{o}$ and $\mathbf{E}_{r}$) to zero vectors.

\subsection{Model Training}

\subsubsection{Training Strategy}

We adopt a two-stage training strategy inspired by visual instruction tuning \citep{liu2023visual}. In the first stage, 3DCity-LLM model is trained on simple caption tasks from the 3DCity-LLM-1.2M dataset, establishing feature alignment across text, 2D image, and 3D point cloud modalities. In the second stage, 3DCity-LLM model is fine-tuned on the high-level analysis and planning tasks from the 3DCity-LLM-1.2M dataset, enabling the model to achieve robust perception and understanding in complex city environments and support object-level, relationship-level and scene-level downstream tasks.

\subsubsection{Loss Function}

3DCity-LLM optimizes its trainable parameters $\theta$ to minimize the negative log-likelihood of generating the target response given the input language instruction and multi-modality context. Formally, for each training pair $(s^{\text{input}}, s^{\text{target}})$, the objective is defined as the token-level cross-entropy loss:

\begin{equation}
\mathcal{L}(\theta) = 
- \frac{1}{n} 
\sum_{i=1}^{n} 
\log P_{\theta}\!\left(
s^{\text{target}}_{i} \mid s^{\text{input}}, s^{\text{target}}_{<i}
\right),
\end{equation}

where $s^{\text{input}}$ denotes the instruction sequence together with the object embeddings, relationship embeddings and scene embeddings, $s^{\text{target}}_{<i}$ represents the previously generated $i{-}1$ tokens in the response, and $n$ is the total number of tokens in $s^{\text{target}}$. The parameter set $\theta$ comprises the feature projectors and all learnable components of the LLM backbone.

\section{Experiments}\label{sec5}

In this section, we first describe the experimental settings including the baseline methods, datasets, and evaluation metrics used in our experiments. Then, we report the experimental results of the proposed 3DCity-LLM in comparison with competitive baseline methods. Finally, we present ablation studies, generalization experiments and fairness verification to further demonstrate the effectiveness of 3DCity-LLM.

\subsection{Experimental Settings}

\subsubsection{Baselines}

We compare 3DCity-LLM with a diverse set of baselines covering two major categories: training-required MLLMs (LLaVA \citep{liu2023visual}, Chat 3D \citep{wang2023chat}, Chat Scene \citep{huang2024chat}, City-VLM \citep{sun2025city}) and training-free large vision-language models (LVLMs) (e.g., ChatGPT-5.1 \citep{achiam2023gpt}, Qwen3-VL Max \citep{yang2025qwen3}, DeepSeek-R1 \citep{guo2025deepseek} and Gemini 3 \citep{comanici2025gemini}). For a fair comparison, all training-required baselines are built upon 7B-parameter LLM backbones, following the same configuration as 3DCity-LLM. In contrast, the training-free LVLMs with substantially higher parameter scales operate directly in zero-shot inference mode without any task-specific adaptation.

\textbf{LLaVA} aligns visual features from a pre-trained CLIP encoder with the embedding space of an LLM through visual instruction tuning. By fine-tuning on image–text conversation data, it enables open-ended visual understanding and multi-modality dialogue.

\textbf{Chat 3D} integrates the perceptual capability of pre-trained 3D encoders with the conversational reasoning of LLMs, forming one of the earliest unified dialogue frameworks for interactive understanding of 3D scenes.

\textbf{Chat Scene} encodes each scene as a sequence of object-level embeddings derived from semantically enriched 2D and 3D features, enabling fine-grained reasoning and strong performance across both object-centric and holistic scene understanding tasks.

\textbf{City-VLM} targets large-scale outdoor scene understanding by introducing an incomplete multi-modality learning framework that models cross-modal interactions via a joint probabilistic distribution space, surpassing prior LVLMs by a large margin on outdoor question-answering benchmarks.

\textbf{ChatGPT-5.1, Qwen3-VL Max, DeepSeek-R1 and Gemini 3} represent the new generation of LVLMs that unify multi-modality understanding, reasoning, and generation within a single framework. Built upon powerful language backbones exceeding hundreds of billions of parameters, they integrate visual and linguistic modalities through large-scale pre-training on diverse corpora encompassing image–text pairs, web documents, and human-aligned conversational data. In this work, we evaluate these state-of-the-art LVLMs via official APIs.

\subsubsection{Datasets}

To demonstrate the effectiveness of our proposed 3DCity-LLM, we adopt the  3DCity-LLM-1.2M dataset and City-3DQA dataset \citep{sun20243d} for evaluation.

\textbf{3DCity-LLM-1.2M} is the self-constructed dataset to advance 3D vision-language in city-scale scenes. We use SensatUrban (Birmingham and Cambridge, UK) \citep{hu2022sensaturban}, UrbanBIS (Qingdao, Lihu, Longhua, Yuehai, and Wuhu, China) \citep{yang2023urbanbis} and City-BIS (Heidelberg, Germany) \citep{li2026city} datasets and generate QA pairs for them. In total, the dataset contains about  1.2 million samples across object caption, object localization, object analysis, relationship computation, scene caption, scene analysis and scene planning tasks.

\textbf{City-3DQA} is a 3D city-scale question-answering benchmark dataset, designed to evaluate fine-grained scene understanding and reasoning in urban environments. City-3DQA dataset covers diverse urban reasoning tasks for scene semantic understanding and human-environment interaction, providing a challenging benchmark for evaluating 3D vision-language models in complex city environments.

\begin{table*}[ht]
\renewcommand{\arraystretch}{1.2}
\centering
\caption{Overall performance comparison on 3DCity-LLM-1.2M dataset. "B-4", "ROU.", "MET.", "Log." and "Rel." are abbreviations for "BLEU-4", "ROUGE-L", "METEOR", "Logicality" and "Reliability", respectively. The score range for "Logicality" and "Reliability" metrics is from 0 to 10. Object-level tasks consist of object caption, object localization and object analysis. Scene-level tasks consist of scene caption, scene analysis and scene planning
}
\resizebox{1.0\linewidth}{!}{
\begin{tabular}{cccccccccccccccc}
\toprule[0.5mm]
\multirow{2}{*}{Method} & \multicolumn{5}{c}{\textit{Object-level Tasks}}  & \multicolumn{5}{c}{\textit{Relationship-level Tasks}} & \multicolumn{5}{c}{\textit{Scene-level Tasks}} 

\\
\cmidrule(lr){2-6} 
\cmidrule(lr){7-11}
\cmidrule(lr){12-16}

& B-4 & ROU. & MET. & Log. & Rel. & B-4 & ROU. & MET. & Log. & Rel. & B-4 & ROU. & MET. & Log. & Rel.
\\ \midrule[0.5mm]

\multicolumn{16}{l}{\textit{\textbf{Training-free but payment-required LVLMs ($>$70B)}}} \\ 

ChatGPT-5.1 & \textbf{17.72} & \textbf{29.01} & 33.95 & \textbf{7.92} & \textbf{6.50} & 16.64 & 30.11 & 38.42 & 6.08 & 4.95 & \textbf{11.44} & \textbf{24.26} & \textbf{35.63} & 7.27 & 6.04 \\

Qwen3-VL Max & 15.23 & 27.12 & 34.65 & 7.05 & 6.39 & \textbf{18.99} & \textbf{32.70} & \textbf{40.69} & \textbf{6.30} & \textbf{5.37} & 9.32 & 21.63 & 32.37 & 7.40 & \textbf{6.12} \\

DeepSeek-R1 & 15.60 & 24.20 & \textbf{37.44} & 7.48 & 6.17 & 14.09 & 27.04 & 37.91 & 5.75 & 4.28 & 10.42 & 20.95 & 31.90 & 6.94 & 5.87 \\

Gemini 3 & 14.94 & 25.02 & 33.87 & 7.45 & 6.47 & 18.06 & 29.50 & 39.52 & 5.62 & 5.07 & 10.60 & 21.91 & 31.33 & \textbf{7.43} & 5.90 \\

\midrule[0.5mm]

\multicolumn{16}{l}{\textit{\textbf{Training-required MLLMs (7B)}}} \\ 

LLaVA & 25.40 & 37.39 & 44.17 & 7.00 & 5.08 & 13.98 & 27.66 & 33.92 & 4.33 & 3.51 & 17.31 & 26.40 & 33.92 & 7.01 & 5.76 \\

Chat 3D & 22.24 & 32.90 & 38.20 & 6.62 & 4.75 & 13.71 & 25.11 & 34.42 & 4.80 & 3.02 & 14.75 & 23.38 & 29.69 & 6.73 & 4.65 \\

Chat Scene & 28.67 & 41.87 & 47.82 & 7.17 & 5.86 & 17.51 & 31.60 & 39.94 & 4.75 & 3.87 & 17.98 & 26.45 & 34.91 & 7.19 & 5.92  \\

City-VLM & 27.93 & 42.13 & 46.84 & 6.96 & 5.58 & 20.48 & \textbf{35.52} & 40.19 & 5.10 & 4.02 & 18.58 & 28.43 & 35.27 & 7.29 & 5.93 \\ 

3DCity-LLM (Ours)  & \textbf{30.64} & \textbf{42.81} & \textbf{48.89} & \textbf{7.33} & \textbf{6.02} & \textbf{20.98} & 34.61 & \textbf{42.09} & \textbf{5.13} & \textbf{4.36} & \textbf{20.11} & \textbf{29.87} & \textbf{37.84} & \textbf{7.50} & \textbf{6.16}

\\ \bottomrule[0.5mm]
\end{tabular}}
\label{table5-OverallTasks}
\vspace{-0.2cm}
\end{table*}

\begin{table*}[t]
\centering
\caption{Overall performance comparison on City-3DQA dataset under sentence-wise and city-wise evaluation settings}
\resizebox{0.75\linewidth}{!}{
\label{table5-City3DQA-OverallTasks}
\renewcommand{\arraystretch}{1.2}
\begin{tabular}{c c c c c c c}
\toprule
\multirow{2}{*}{Method}
& \multicolumn{3}{c}{\textit{Sentence-wise}} 
& \multicolumn{3}{c}{\textit{City-wise}} \\
\cmidrule(lr){2-4} \cmidrule(lr){5-7}
& Single-hop & Multi-hop & All 
& Single-hop & Multi-hop & All \\
\midrule
ScanQA~\citep{azuma2022scanqa}        & 76.42 & 28.31 & 49.28 & 64.84 & 27.03 & 47.33 \\
3D-VisTA~\citep{zhu20233d}     & 79.23 & 44.67 & 59.63 & 71.28 & 43.87 & 56.74 \\
Sg-CityU~\citep{sun20243d}  & 80.95 & 50.75 & 63.94 & 78.46 & 50.50 & 63.76 \\
City-VLM~\citep{sun2025city}  & 81.74 & 56.80 & 67.30 & 78.84 & \textbf{52.26} & \textbf{64.70} \\
3DCity-LLM (Ours)  & \textbf{82.41} & \textbf{57.75} & \textbf{68.55} & \textbf{79.10} & 51.23  & 64.49 \\
\bottomrule
\end{tabular}}
\end{table*}

\subsubsection{Evaluation Metrics} 

To evaluate the model outputs across diverse task types, we adopt both text-based similarity metrics and LLM-based comprehensive metrics.

\textit{a) Text-based Similarity Metrics.} For tasks with textual outputs, we follow established evaluation protocols \citep{wang2023chat, huang2024chat} and report BLEU, ROUGE-L and METEOR as primary metrics.

\textit{b) LLM-based Comprehensive Metrics.} Text-based similarity metrics often struggle to assess long-form answers, especially when their meaning is preserved despite variations in wording or sentence structure. Inspired by recent works \citep{qi2025gpt4scene, linghu2024multi}, we also report two comprehensive LLM-based metrics: Logicality and Reliability.

$\bullet$ Logicality measures the internal coherence and reasoning validity of a generated response. A logically strong answer should demonstrate a clear argumentative flow, avoid contradictions, and present conclusions that follow consistently from the given premises, regardless of lexical overlap with the reference.

$\bullet$ Reliability indicates the factual correctness and evidence alignment of a response. A reliable answer must faithfully reflect the provided ground-truth information and scene evidence, avoiding hallucinated details or unsupported claims, thereby ensuring trustworthiness.

To obtain these comprehensive scores, we employ three LLMs including ChatGPT-5 \citep{achiam2023gpt}, Qwen3-VL Plus \citep{yang2025qwen3} and DeepSeek-V3 \citep{liu2024deepseek} as the independent evaluators. Each LLM is presented with the generated answer, the ground truth, and the relevant 3D scene evidence, without disclosure of the source model. Importantly, each LLM required to rate the generated answers for the Logicality and Reliability metrics on a scale from 0 to 10, and provide a concise justification to ensure that every score is accompanied by an explicit and traceable rationale. The final Logicality and Reliability metrics is calculated as the average across evaluators.

\subsubsection{Implementation Details}

We build the 3DCity-LLM model on the pre-trained \texttt{LLaVA-v1.5-7B} \citep{liu2023visual}. For model fine-tuning, we adopt LoRA \citep{hu2022lora} on the attention layers, modality projectors, and feed-forward components. The LoRA rank is set to 8, with a dropout rate of 0.05. The training process is performed with a maximum sequence length of 512 tokens using brain floating point 16-bit (bf16) precision, a batch size of 8, and 10-step gradient accumulation. We use the AdamW optimizer with a learning rate of $5 \times 10^{-4}$ for model optimization. A warm-up schedule is applied for the first 10 steps before decaying the learning rate, and DeepSpeed ZeRO Stage 2 optimization is employed to reduce memory usage. All the experiments are implemented with PyTorch on a single NVIDIA A100 GPU. The training process of the first and second stage takes about 10 and 24 hours to converge, respectively.

\begin{table*}[ht]
\renewcommand{\arraystretch}{1.2}
\centering
\caption{Performance comparison on object-level tasks in 3DCity-LLM-1.2M dataset. "B-4", "ROU.", "MET.", "Log." and "Rel." are abbreviations for "BLEU-4", "ROUGE-L", "METEOR", "Logicality" and "Reliability", respectively. The score range for "Logicality" and "Reliability" metrics is from 0 to 10
}
\resizebox{1.0\linewidth}{!}{
\begin{tabular}{cccccccccccccccc}
\toprule[0.5mm]
\multirow{2}{*}{Method} & \multicolumn{5}{c}{\textit{Object Caption}}  & \multicolumn{5}{c}{\textit{Object Localization}} & \multicolumn{5}{c}{\textit{Object Analysis}} 

\\
\cmidrule(lr){2-6} 
\cmidrule(lr){7-11}
\cmidrule(lr){12-16}

& B-4 & ROU. & MET. & Log. & Rel. & B-4 & ROU. & MET. & Log. & Rel. & B-4 & ROU. & MET. & Log. & Rel.
\\ \midrule[0.5mm]

LLaVA & 11.98 & 22.01 & 29.52 & 5.32 & 3.92 & 14.75 & 28.19 & 31.10 & 5.98 & 3.96 & 32.57 & 45.01  & 52.29 & 7.84 & 6.46 \\

Chat 3D & 9.75 & 18.55 & 25.74 & 5.10 & 2.98 & 11.91 & 23.97 & 26.41 & 3.64 & 3.07 & 28.99 & 40.07 & 45.24 & 7.79 & 5.75 \\

Chat Scene & 12.29 & 22.70 & 30.95 & 5.62 & 4.20 & 19.01 & 32.36 & 35.50 & 5.91 & 4.10 & 36.74 & 50.96 & 56.62 & 8.01 & 6.84 \\

City-VLM & 11.75 & 20.49 & 31.97 & 5.85 & 4.01 & \textbf{20.07} & \textbf{33.23} & 37.20 & 6.12 & 4.34 & 35.57 & 52.02 & 54.36 & 7.56 & 6.43 \\

3DCity-LLM (Ours) & \textbf{13.80} & \textbf{23.24} & \textbf{32.25} & \textbf{5.89} & \textbf{4.37} & 19.36 & 30.14 & \textbf{39.41} & \textbf{6.28} & \textbf{4.52} & \textbf{39.22} & \textbf{52.69} & \textbf{57.02} & \textbf{8.09} & \textbf{6.95}

\\ \bottomrule[0.5mm]
\end{tabular}}
\label{table6-ObjectTask}
\end{table*}

\begin{table*}[ht]
\renewcommand{\arraystretch}{1.2}
\centering
\caption{Performance comparison on scene-level tasks in 3DCity-LLM-1.2M dataset. "B-4", "ROU.", "MET.", "Log." and "Rel." are abbreviations for "BLEU-4", "ROUGE-L", "METEOR", "Logicality" and "Reliability", respectively. The score range for "Logicality" and "Reliability" metrics is from 0 to 10
}
\resizebox{1.0\linewidth}{!}{
\begin{tabular}{cccccccccccccccc}
\toprule[0.5mm]
\multirow{2}{*}{Method} & \multicolumn{5}{c}{\textit{Scene Caption}}  & \multicolumn{5}{c}{\textit{Scene Analysis}} & \multicolumn{5}{c}{\textit{Scene Planning}} 

\\
\cmidrule(lr){2-6} 
\cmidrule(lr){7-11}
\cmidrule(lr){12-16}

& B-4 & ROU. & MET. & Log. & Rel. & B-4 & ROU. & MET. & Log. & Rel. & B-4 & ROU. & MET. & Log. & Rel.
\\ \midrule[0.5mm]

LLaVA & 20.74 & 26.19 & 30.37 & 7.14 & 6.03 & 14.82 & 24.20 & 32.93 & 6.96 & 5.04 & 15.57 & 28.41 & 38.56 & 6.91 & 6.06 \\

Chat 3D & 18.46 & 24.81 & 26.39 & 6.82 & 4.19 & 12.09 & 20.07 & 29.83 & 6.58 & 4.73 & 12.84 & 24.48 & 33.20 & 6.76 & 5.09 \\

Chat Scene & 23.44 & 28.90 & 32.98 & 6.78 & 6.27 & 14.30 & 23.39 & 34.09 & 7.09 & 5.50 & 14.97 & 26.25 & 37.69 & \textbf{7.72} & 5.85 \\

City-VLM & 22.51 & 30.20 & 31.54 & 7.12 & 5.94 & 15.79 & 24.94 & 34.27 & 7.24 & 5.44 & 16.54 & \textbf{29.32} & 40.15 & 7.54 & \textbf{6.30} \\

3DCity-LLM (Ours) & \textbf{25.01} & \textbf{31.75} & \textbf{35.08} & \textbf{7.65} & \textbf{6.50} & \textbf{17.42} & \textbf{28.54} & \textbf{36.60} & \textbf{7.30} & \textbf{5.76} & \textbf{16.92} &  28.87 & \textbf{41.85} & 7.51 & 6.12

\\ \bottomrule[0.5mm]
\end{tabular}}
\label{table7-SceneTask}
\end{table*}

\subsection{Quantitative Results}

\textbf{Comparison among LVLMs in 3DCity-LLM-1.2M dataset.} As shown in the upper half of Table \ref{table5-OverallTasks}, we compare training-free but payment-required LVLMs, including ChatGPT-5.1, Qwen3-VL Max, DeepSeek-R1, and Gemini 3. Across all three task levels, LVLMs exhibit strong zero-shot performances, particularly on object-level tasks. For instance, ChatGPT-5.1 achieves BLEU-4 of 17.72 and 11.44, ROUGE-L of 29.01 and 24.26 on object-level and scene-level tasks, respectively. Qwen3-VL Max attains the highest BLEU-4 of 18.99, ROUGE-L of 32.70, METEOR of 40.69 on relationship-level tasks. In contrast, DeepSeek-R1 and Gemini 3 demonstrate comparable but slightly lower performances. Notably, although LVLMs achieve relatively lower scores on text-similarity metrics, they perform well on LLM-based evaluations of logicality and reliability. This indicates that LVLMs tend to produce semantically coherent and well-reasoned responses rather than strict lexical alignment with ground-truth answers.

\begin{figure*}[]
\centering
\includegraphics[width=1.0\textwidth]{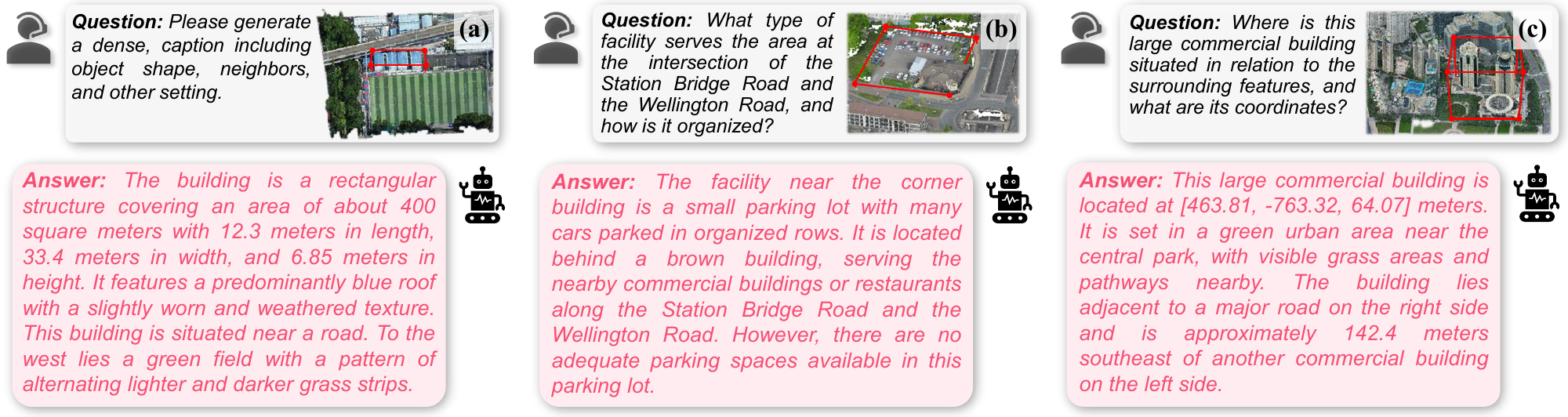}
\vspace{-0.3cm}
\caption{Qualitative results on the object-level tasks from 3DCity-LLM-1.2M dataset. (a) Object Caption, (b) Object Analysis, (c) Object Localization}
\label{Fig-CityLLMObjectTask}
\end{figure*}

\begin{figure*}[]
\centering
\includegraphics[width=1.0\textwidth]{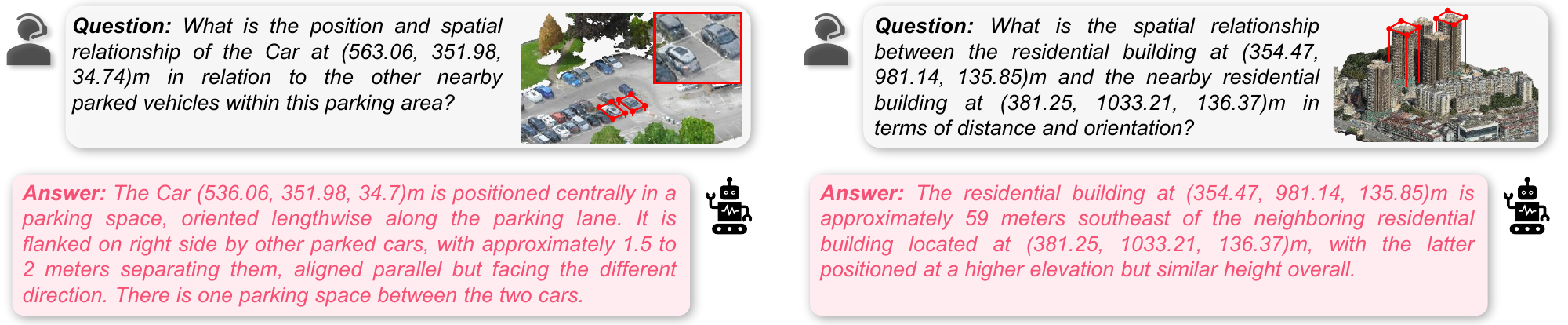}
\vspace{-0.3cm}
\caption{Qualitative results on the relationship-level tasks from 3DCity-LLM-1.2M dataset. 3DCity-LLM model can extract object information in urban scenes and calculate distance and orientation based on their coordinates}
\label{Fig-CityLLMRelationshipTask}
\end{figure*}

\textbf{Comparison among MLLMs in 3DCity-LLM-1.2M dataset.} As shown in the bottom half of Table \ref{table5-OverallTasks}, we analyze training-required MLLMs, including LLaVA, Chat 3D, Chat Scene, City-VLM, and our proposed 3DCity-LLM. Compared with LLaVA, models explicitly designed for scene understanding (e.g., Chat Scene and City-VLM) achieve better performance on object-level and relationship-level tasks, underscoring the importance of explicit spatial and structural modeling. For example, Chat Scene improves BLEU-4 from 25.40 (LLaVA) to 28.67 on object-level tasks and achieves higher METEOR of 39.94 on relationship-level tasks. Notably, 3DCity-LLM outperforms all other baseline methods across three task types, obtaining the highest object-level BLEU-4 of 30.64 and reliability of 6.02, as well as superior scene-level performance with a BLEU-4 of 20.11 and METEOR of 37.84.

\textbf{Comparison among different methods in City-3DQA dataset.} Table \ref{table5-City3DQA-OverallTasks} reports the performance comparison on the City-3DQA dataset under both sentence-wise and city-wise settings. 3DCity-LLM attains the highest accuracy of 68.55 under sentence-wise setting, outperforming all baseline methods, which demonstrates its strong capability in fine-grained question answering. Under city-wise evaluation, 3DCity-LLM attains the highest single-hop accuracy of 79.10 and achieves the second-best accuracy of 64.49, closely following City-VLM method with the best accuracy of 64.70.

\textbf{Comparison on sub-tasks in 3DCity-LLM-1.2M dataset.} Table~\ref{table6-ObjectTask} presents the performance comparison on object-level tasks, including object caption, object localization, and object analysis. 3DCity-LLM achieves the highest BLEU-4 and ROUGE-L metrics on object caption, with BLEU-4 reaching 13.80 and ROUGE-L reaching 23.24, and attains the best performance on object analysis, achieving BLEU-4 of 39.22 and METEOR of 57.02. These results demonstrate the effectiveness of the fine-grained object feature encoding strategy in 3DCity-LLM for urban object perception.
Table~\ref{table7-SceneTask} reports the results on scene-level tasks, including scene caption, scene analysis, and scene planning, which require holistic perception and long-range spatial reasoning in urban environments. 3DCity-LLM achieves the best overall performance on scene captioning and scene analysis under both text-similarity metrics and LLM-based comprehensive metrics, attaining the highest BLEU-4 of 25.01 for scene caption and 17.42 for scene analysis, outperforming prior scene-aware methods such as Chat Scene and City-VLM. For the more challenging scene planning task, which involves decision-making and action trade-offs, 3DCity-LLM achieves the highest logicality of 7.51 and reliability of 6.12, while maintaining competitive text-based performance with the BLEU-4 of 16.92 and METEOR of 41.85.

\subsection{Qualitative Visualization}

\begin{table}[t]
\centering
\caption{Ablation study of different feature encoding for object-level tasks}
\label{tab:feature_embedding_object_task}
\fontsize{7pt}{10pt}\selectfont
\setlength{\tabcolsep}{2pt}
\begin{tabular}{cccccccc}
\hline
\multicolumn{3}{c}{Feature Encoding} & \multicolumn{5}{c}{\textit{Object-level Tasks}} \\
\cmidrule(lr){1-3}
\cmidrule(lr){4-8}
Local View & Shape & Landmark & B-4 & ROU. & MET. & Log. & Rel. \\
\hline
$\checkmark$ &  &  & 29.61 & 42.66 & 45.94 & 7.25 & 5.64 \\
$\checkmark$ & $\checkmark$ &  & 29.82 & 42.47 & 46.50 & 7.20 & 5.72 \\
$\checkmark$ & $\checkmark$ & $\checkmark$ & 30.64 & 42.81 & 48.89 & 7.33 & 6.02 \\
\hline
\end{tabular}
\end{table}

The qualitative visualizations on the object-level tasks are shown in Fig. \ref{Fig-CityLLMObjectTask}. For the object caption and object analysis tasks, 3DCity-LLM produces well-structured and information-rich descriptions of urban objects. The generated answers consistently capture geometric attributes such as shape, scale, and roof type, while also integrating semantic properties and surrounding contextual cues, including adjacent roads, vegetation layouts, and nearby facilities.

\begin{table}[t]
\centering
\caption{Ablation study of different feature encoding for relationship-level tasks}
\fontsize{8pt}{12pt}\selectfont
\setlength{\tabcolsep}{3pt}
\begin{tabular}{ccccccc}
\hline
\multicolumn{2}{c}{Feature Encoding} & \multicolumn{5}{c}{\textit{Relationship-level Tasks}} \\
\cmidrule(lr){1-2} 
\cmidrule(lr){3-7}
Geometry & Landmark & B-4 & ROU. & MET. & Log. & Rel. \\
\hline
$\checkmark$ &  & 18.24 & 33.87 & 41.26 & 5.01 & 4.04   \\
$\checkmark$ & $\checkmark$ & 20.98 & 34.61 & 42.09 & 5.13 & 4.36   \\
\hline
\end{tabular}
\label{tab:feature_embedding_relationship_task}
\end{table}

As illustrated in Fig. \ref{Fig-CityLLMRelationshipTask}, for the relationship computation task, 3DCity-LLM can accurately infer relative orientation and distance between objects based on their absolute coordinates in city scenes. In addition to quantitative spatial reasoning, 3DCity-LLM provides coherent relational descriptions by referencing nearby buildings and road structures, reflecting a comprehensive understanding of local spatial configurations.

\begin{figure*}[]
\centering
\includegraphics[width=0.98\textwidth]{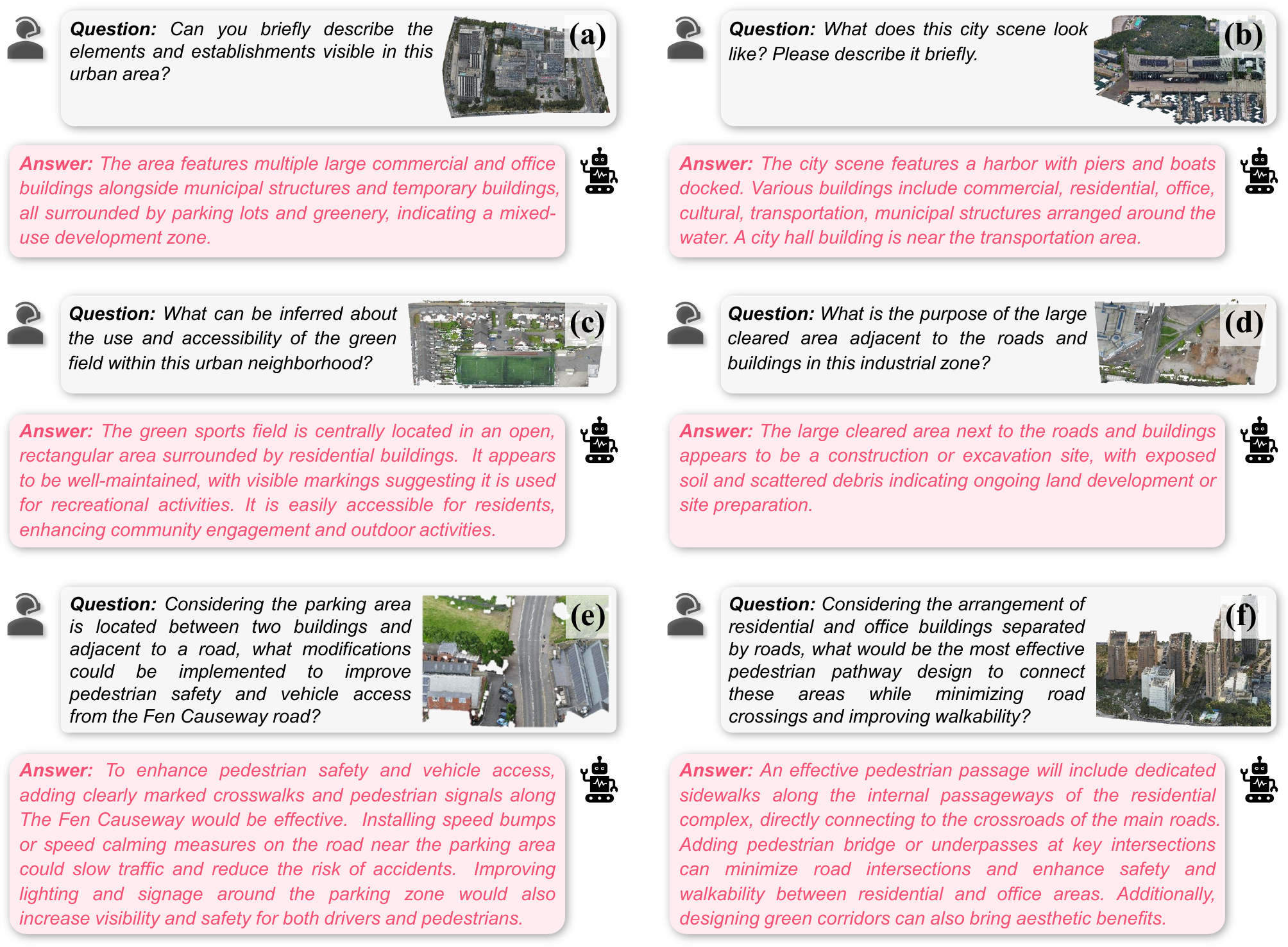}
\caption{Qualitative results on the scene-level tasks from 3DCity-LLM-1.2M dataset. (a-b) Scene Caption, (c-d) Scene Analysis, (e-f) Scene Planning}
\label{Fig-CityLLMSceneTask}
\end{figure*}

\begin{table}[t]
\centering
\caption{Ablation study of different feature encoding for scene-level tasks}
\fontsize{8pt}{12pt}\selectfont
\setlength{\tabcolsep}{2.6pt}
\begin{tabular}{ccccccc}
\hline
\multicolumn{2}{c}{Feature Encoding} & \multicolumn{5}{c}{\textit{Scene-level Tasks}} \\
\cmidrule(lr){1-2} 
\cmidrule(lr){3-7}
Global View & Landmark & B-4 & ROU. & MET. & Log. & Rel. \\
\hline
$\checkmark$ & & 19.72 & 29.03 & 35.82 & 7.47 & 5.93    \\
$\checkmark$ & $\checkmark$ & 20.11 & 29.87 & 37.84 & 7.50 & 6.16   \\
\hline
\end{tabular}
\label{tab:feature_embedding_scene_task}
\end{table}

3DCity-LLM also demonstrates consistent and robust performance for holistic urban understanding. For the scene caption task shown in Fig. \ref{Fig-CityLLMSceneTask} (a–b), 3DCity-LLM is able to generate concise but comprehensive summaries of complex urban environments, and accurately identifies dominant functional zones, such as commercial districts, residential areas and transportation infrastructure. These scene captions reflect an integrated understanding of object distributions, land-use patterns, and spatial layouts. In addition to surface-level recognition, 3DCity-LLM can provide higher-level interpretations grounded in the structural and contextual attributes of the city scene. For the scene analysis task shown in Fig.~\ref{Fig-CityLLMSceneTask} (c–d), 3DCity-LLM reasons about the usage of open spaces, green fields, and cleared areas by jointly considering surrounding buildings, road networks, and spatial openness. For the scene planning task shown in Fig.~\ref{Fig-CityLLMSceneTask} (e–f), 3DCity-LLM exhibits goal-oriented reasoning by proposing feasible modifications to improve pedestrian safety, accessibility, and urban connectivity. The generated suggestions, such as introducing pedestrian crossings, traffic-calming measures, dedicated walkways, or green corridors, are spatially grounded and aligned with the given scene layout.

\subsection{Ablation Study}

Tables \ref{tab:feature_embedding_object_task}–\ref{tab:feature_embedding_scene_task} present ablation results evaluating different feature encoding configurations for object-level, relationship-level, and scene-level tasks.

\begin{figure*}[t]
\centering
\includegraphics[width=1.00\textwidth]{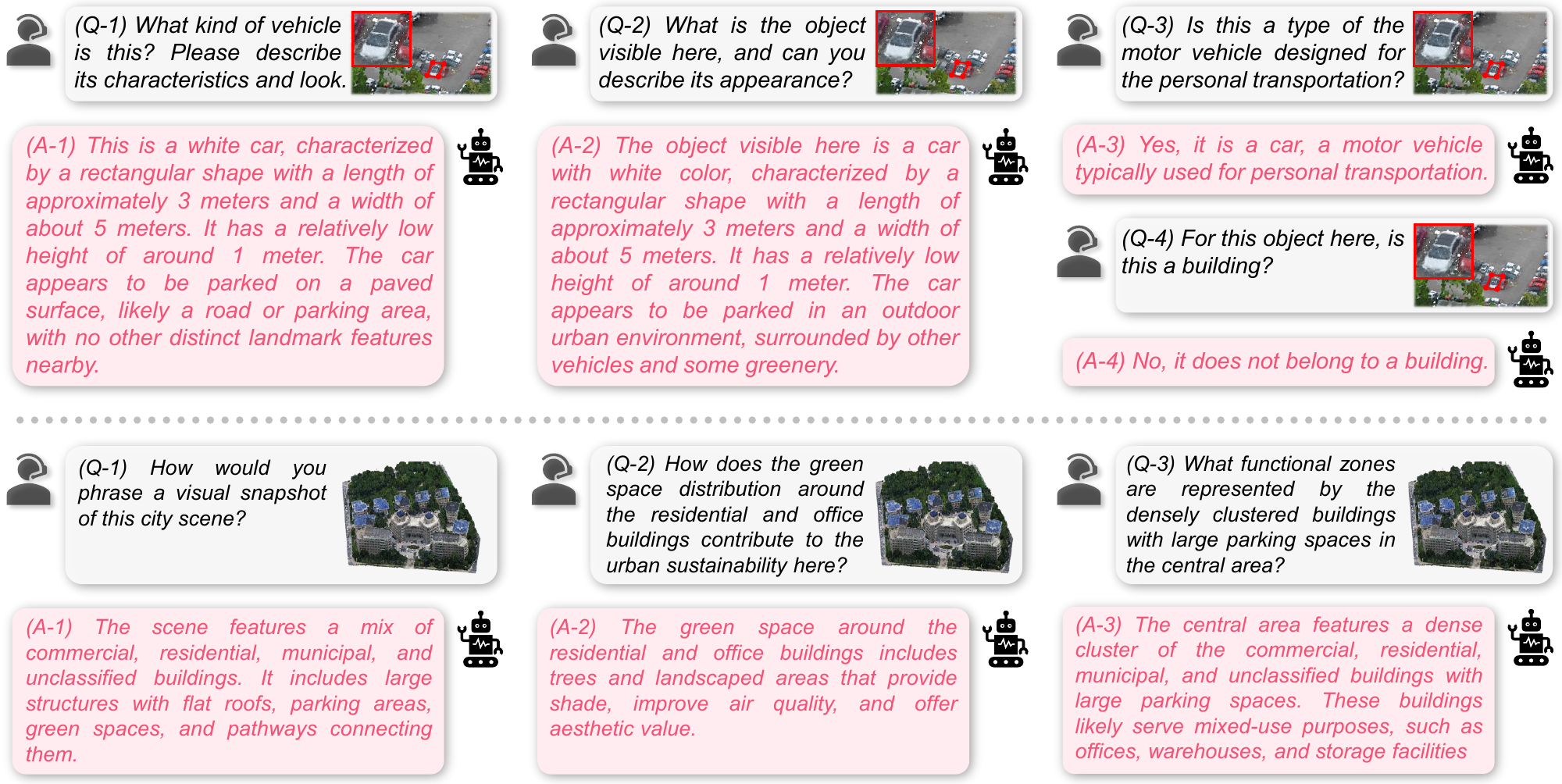}
\vspace{-0.4cm}
\caption{Qualitative examples of diverse queries for the same object (top) or scene (bottom). Queries with varying linguistic styles and word lengths are employed to evaluate the 3DCity-LLM's generalization ability on text modality}
\vspace{-0.3cm}
\label{Fig-CityLLMDiverseQueries}
\end{figure*}

\begin{figure*}[t]
\centering
\includegraphics[width=1.00\textwidth]{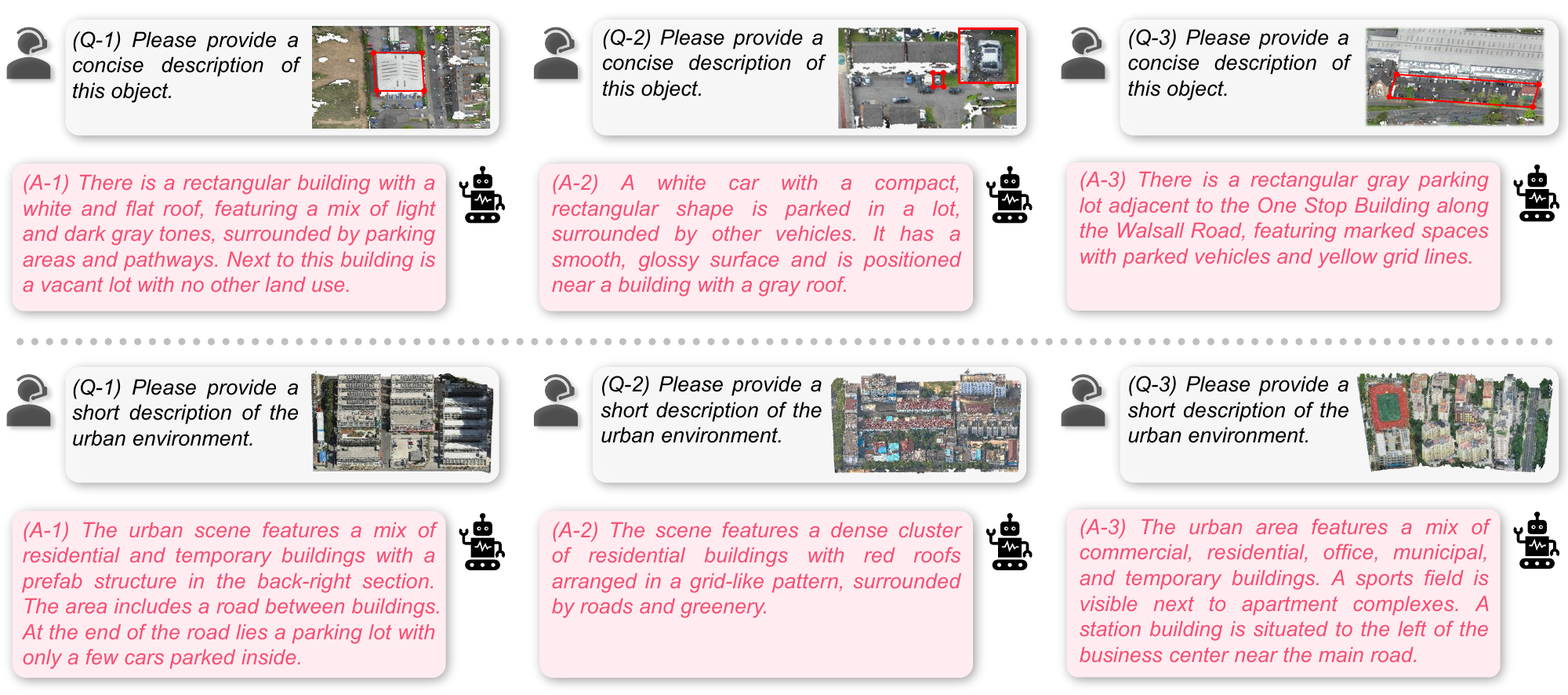}
\vspace{-0.4cm}
\caption{Qualitative examples of the same queries for diverse object (top) or scene (bottom). Different objects and scenes are employed to evaluate the 3DCity-LLM's generalization ability on 3D modality}
\label{Fig-CityLLMDiverseObjectScene}
\end{figure*}

\begin{figure*}[t]
\centering
\includegraphics[width=1.0\textwidth]{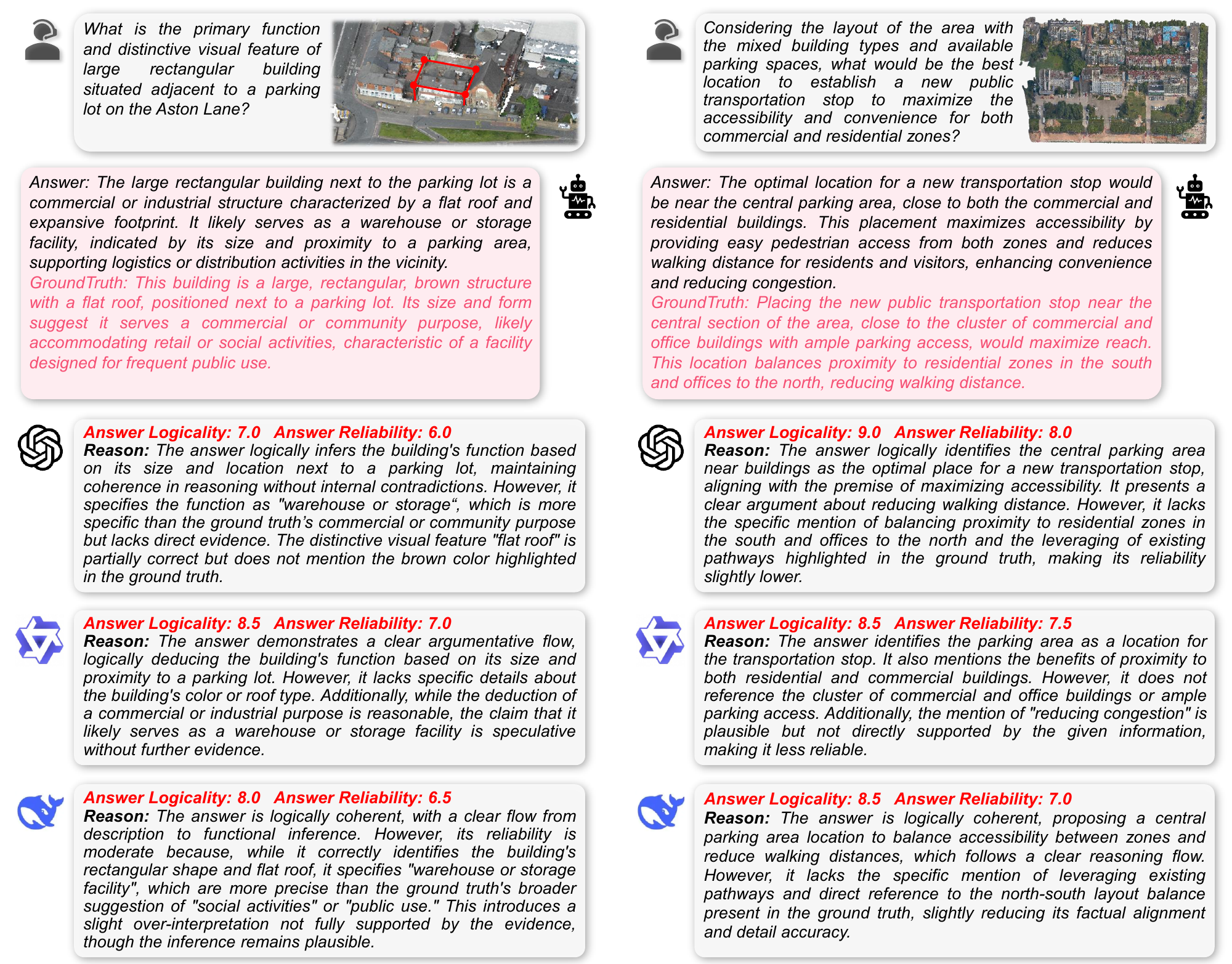}
\vspace{-0.4cm}
\caption{The representative examples of LLM-based evaluation process from three LLM evaluators (ChatGPT-5, Qwen3-VL Plus and DeepSeek-V3). Each evaluator assigns logicality and reliability metrics with explicit scoring justification for evidence usage, specificity, and spatial-reasoning completeness}
\label{Fig-CityLLMScoreProof}
\end{figure*}

The quantitative ablation results for object-level tasks are shown in Table \ref{tab:feature_embedding_object_task}, encoding only the local view feature provides a strong baseline, achieving a BLEU-4 metric of 29.61 and a METEOR metric of 45.94. Incorporating shape information yields marginal gains, while jointly encoding shape and landmark features leads to more noticeable improvements. Specifically, BLEU-4 increases by 1.03 and METEOR improves by 2.95 compared to using the local view feature alone. 

As shown in Table \ref{tab:feature_embedding_relationship_task}, geometry-only encoding achieves limited performance with BLEU-4 of 18.24 and reliability metric of 4.04 for relationship-level tasks. Introducing landmark features improves BLEU-4 by 2.74 and reliability by 0.32. 

For the ablation results of scene-level tasks shown in Table \ref{tab:feature_embedding_scene_task}, when only the global view feature is encoded, 3DCity-LLM achieves a solid baseline performance with BLEU-4 of 19.72, ROUGE-L of 29.03 and METEOR of 35.82, indicating that top-view maps are effective in capturing the overall spatial layout and structural composition of city scenes. Introducing landmark features provide explicit real-world geographic references for well-known urban objects, leading to consistent improvements across all metrics. BLEU-4 increases from 19.72 to 20.11, ROUGE improves from 29.03 to 29.87, and METEOR rises from 35.82 to 37.84. In addition, gains in llm-based metrics (logicality and reliability) reflect more coherent reasoning and closer alignment with human cognitive patterns.

\begin{table*}[t]
\centering
\renewcommand{\arraystretch}{1.2}
\caption{Comparison of logicality and reliability metrics provided by three independent LLM evaluators (ChatGPT-5, Qwen3-VL Plus, DeepSeek-V3) for all seven task categories in 3DCity-LLM-1.2M dataset. By averaging the logicality and reliability metrics from different evaluators, the final assessment is more stable and less affected by evaluator-specific biases}
\resizebox{1.00\linewidth}{!}{
\begin{tabular}{lcccccccc}
\toprule
\multirow{2}{*}{Task Category} 
& \multicolumn{4}{c}{\textit{Logicality}} 
& \multicolumn{4}{c}{\textit{Reliability}} \\
\cmidrule(lr){2-5} \cmidrule(lr){6-9}
& ChatGPT-5 & Qwen3-VL Plus & DeepSeek-V3 & Average 
& ChatGPT-5 & Qwen3-VL Plus & DeepSeek-V3 & Average \\
\midrule
Object Caption        & 5.87 & 5.74 & 6.08 & 5.89 & 4.75 & 3.88 & 4.50 & 4.37 \\
Object Localization   & 6.56 & 6.65 & 5.62 & 6.28 & 4.06 & 5.79 & 3.72 & 4.52 \\
Object Analysis        & 8.14 & 8.28 & 7.85 & 8.09 & 7.06 & 6.98 & 6.81 & 6.95 \\
Relationship Computation  & 4.59 & 6.46 & 4.34 & 5.13 & 4.40 & 5.58 & 3.12 & 4.36 \\
Scene Caption         & 7.61 & 7.39 & 7.95 & 7.65 & 6.59 & 6.05 & 6.87 & 6.50 \\
Scene Analysis        & 7.44 & 7.16 & 7.31 & 7.30 & 6.04 & 5.67 & 5.59 & 5.76 \\
Scene Planning        & 7.45 & 7.87 & 7.21 & 7.51 & 6.29 & 6.46 & 5.59 & 6.11 \\
\bottomrule
\end{tabular}}
\label{tab:llm_evaluation}
\end{table*}

\begin{figure*}[t]
\centering
\vspace{-0.4cm}
\includegraphics[width=1.0\textwidth]{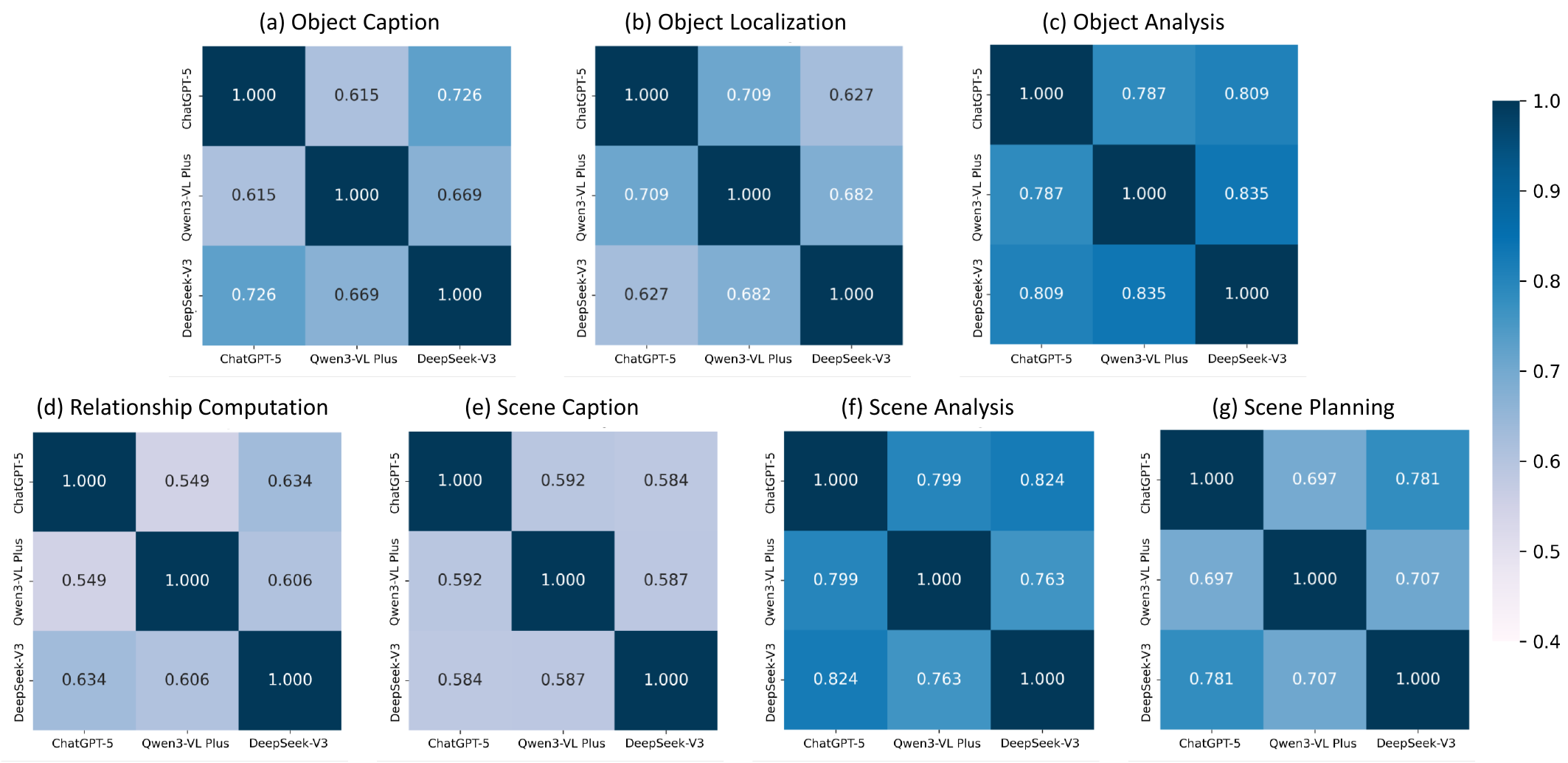}
\vspace{-0.4cm}
\caption{Correlation heatmap of LLM metrics across the object caption, object localization, object analysis, relationship computation, scene caption, scene analysis, and scene planning tasks. The LLM-based metrics show moderately strong but not absolute inter-evaluator correlation with the most values in 0.5 to 0.8}
\label{Fig-CityLLMScoreRel}
\end{figure*}

\subsection{Generalization Experiments}

\textbf{Diverse Queries.} Fig.~\ref{Fig-CityLLMDiverseQueries} presents the qualitative results of diverse query formulations for the same target object or scene. Although the questions differ in wording, emphasis, and inquiry manner, 3DCity-LLM consistently produces accurate and coherent responses.

At the object level, 3DCity-LLM can recognize object categories and extract their attributes such as geometry, scale and visual appearances across different question styles. Importantly, the final answers adapt to the specific focus of each query without internal inconsistency, indicating robust language instruction understanding rather than reliance on fixed query templates. At the scene level, while maintaining stable interpretations of the urban environment, 3DCity-LLM can adjust the emphasis such as functional composition or spatial layout, demonstrating strong linguistic generalization and sensitivity to nuanced changes in user intent.

\textbf{Diverse Objects and Scenes.} Fig. \ref{Fig-CityLLMDiverseObjectScene} illustrates qualitative results of diverse objects and scenes under the same query formulation. Despite significant variations in object categories and scene layouts, 3DCity-LLM can generate the scene-grounded descriptions, thereby conducting logical spatial reasoning and analysis.

For object-level cases, 3DCity-LLM adapts to heterogeneous objects including buildings, vehicles, and parking lots, indicating that 3DCity-LLM learns transferable object representations instead of memorizing query-specific response patterns. For scene-level cases, 3DCity-LLM captures high-level characteristics such as land-use type, building density, road topology, and green space distribution before generating coherent caption or analysis, demonstrating its generalization across diverse objects or scenes.

\subsection{Fairness Verification}

\textbf{Traceability of LLM-based Metrics.} To ensure rating traceability in LLM-based evaluation, we require each LLM evaluator to produce not only a numerical score but also an explicit reasoning trace that justifies its scoring decision. This mechanism prevents opaque assessments and forces evaluators to articulate how the generated answer aligns or conflicts with the observable scene evidences.

As illustrated in Fig. \ref{Fig-CityLLMScoreProof}, LLM evaluators (ChatGPT-5, Qwen3-VL Plus and DeepSeek-V3) explicitly reference visual cues, spatial context, and ground-truth annotations when generating both the logicality and reliability metrics and the corresponding justification. In the case of object analysis shown in the left of Fig. \ref{Fig-CityLLMScoreProof}, all evaluators acknowledge that 3DCity-LLM appropriately grounds its answer in observable evidences (flat roof, large footprint, and proximity to the parking lot). However, evaluators point out that the predicted “warehouse or storage facility” over-specifies the building function compared to the ground-truth (“commercial or community purpose”), resulting in a lower reliability score. In the case of scene planning shown in the right of Fig. \ref{Fig-CityLLMScoreProof}, the LLM evaluators assess the generated answer by checking whether key factors for transportation stop selection (such as accessibility for residential and commercial areas or making use of existing pedestrian routes) are considered.

\textbf{Correlation Analysis for LLM-based Metrics.} As shown in Fig. \ref{Fig-CityLLMScoreRel} and Table \ref{tab:llm_evaluation}, we present the specific LLM-based metrics and visualize the correlation heatmaps for three LLM evaluators. Although all evaluators receive identical inputs (model-generated answer, ground-truth, and structured scene evidence), their scoring behaviors are not absolutely correlated. Most correlation values fall between 0.5 and 0.8, indicating that the evaluators share certain judgment patterns but do not collapse into dominant preference and no single evaluator disproportionately influences the evaluation results.

\subsection{Efficiency Analysis}

\begin{table}[t]
\centering
\caption{Efficiency and performance of 3DCity-LLM. We report BLEU-4, ROUGE-L, METEOR, logicality, reliability, and average inference time}
\fontsize{6pt}{8pt}\selectfont
\setlength{\tabcolsep}{2.2pt}
\begin{tabular}{l c c c c c c}
\toprule
Task Category & B-4 & ROU. & MET. & Log. & Rel. & Time (s) \\
\midrule
Object-level Task & 30.64 & 42.81 & 48.89 & 7.33 & 6.02 & 7.40 \\
Relationship-level Task & 20.98 & 34.61 & 42.09 & 5.13 & 4.36 & 13.69 \\
Scene-level Task & 20.11 & 29.87 & 37.84 & 7.50 & 6.16 & 14.54 \\
\bottomrule
\end{tabular}
\label{tab:Efficiency_Analysis}
\end{table}

Table \ref{tab:Efficiency_Analysis} summarizes the computational efficiency and performance of 3DCity-LLM across the three task categories. 3DCity-LLM achieves the fastest execution on object-level tasks, requiring only 7.40 seconds per inference. In comparison, relationship-level tasks increase the inference time to 13.69 seconds because they require jointly encoding two or more object-centric features and performing spatial computations for their relative distance and orientation. Scene-level tasks exhibit the highest latency at 14.54 seconds per inference, reflecting the increased complexity of processing a wider spatial context and performing multi-stage reasoning steps required for comprehensive scene understanding.

\section{Discussion}\label{sec6}

\subsection{Practicality}

\textbf{Dataset Volume.} The 3DCity-LLM-1.2M dataset is constructed at a substantially larger volume (1.2 millions) compared to the existing 3D–VL benchmarks and emphasizes practical city-scale perception, reasoning, and decision-making challenges that frequently appear in urban studies. For example, the scene analysis task in 3DCity-LLM-1.2M covers a wide spectrum of urban functional, morphological, environmental, societal, infrastructure-oriented, and development-related analysis. The large-scale volume and task realism allows models trained with sufficient supervision and improves their ability for real-world applications.


\textbf{QA Diversity.} The diverse question-answering styles, urban environments, and contextual simulations enable models to learn more realistic and context-aware perception and reasoning patterns, thereby accommodating the requirements from different user groups.

\subsection{Privacy}

Throughout the data construction process, we adhered to stringent privacy-preservation principles. The 3D city datasets used in this work were sourced from publicly available repositories that had already removed sensitive attributes such as faces, license plates, and resident-level details. All text annotations in 3DCity-LLM-1.2M dataset were generated within the depersonalized reconstructed scenes, ensuring that no identifiable information could be introduced. Additionally, we have performed an additional validation in quality control to identify and remove any samples that might inadvertently pose a privacy risk.

\subsection{Limitation}

Due to GPU resource constraints, we were unable to train or fine-tune 3DCity-LLM using larger foundation models (13B or 34B) as backbones. Prior studies \citep{zhao2023survey} indicate that scaling model parameters often leads to notable performance gains, especially in tasks requiring long-range spatial reasoning or multi-hop inference. Therefore, exploring larger MLLMs remains an important direction for future work.


\section{Conclusion}\label{sec7}

In this paper, we proposed 3DCity-LLM, a multi-modality large language model designed for 3D city-scale perception and understanding. 3DCity-LLM adopts a coarse-to-fine feature encoding strategy and a task-driven instruction tuning paradigm, enabling a single model architecture to support object-level, relationship-level, and scene-level downstream tasks within complex city environments. Furthermore, we introduced 3DCity-LLM-1.2M dataset, which encompasses seven representative task types and is grounded in explicit 3D numerical information with diverse user-oriented contextual simulation. Extensive experiments demonstrated that 3DCity-LLM achieves remarkable performances in two benchmark datasets. Our future work will explore larger LLM backbones and more diverse datasets to narrow the gap between current research and real-world applications.

\backmatter








\bibliography{sn-bibliography}

@article{lobry2020rsvqa,
  title={RSVQA: Visual question answering for remote sensing data},
  author={Lobry, Sylvain and Marcos, Diego and Murray, Jesse and Tuia, Devis},
  journal={IEEE Transactions on Geoscience and Remote Sensing},
  volume={58},
  number={12},
  pages={8555--8566},
  year={2020},
  publisher={IEEE}
}

@inproceedings{wang2024earthvqa,
  title={Earthvqa: Towards queryable earth via relational reasoning-based remote sensing visual question answering},
  author={Wang, Junjue and Zheng, Zhuo and Chen, Zihang and Ma, Ailong and Zhong, Yanfei},
  booktitle={Proceedings of the AAAI conference on artificial intelligence},
  volume={38},
  number={6},
  pages={5481--5489},
  year={2024}
}

@article{li2026city,
  title={City-scale building instance segmentation from LiDAR point clouds via structure-aware method},
  author={Li, Jinpeng and Li, Yuan and Chen, Yiping and Fan, Hongchao and Wang, Ruisheng},
  journal={International Journal of Applied Earth Observation and Geoinformation},
  volume={146},
  pages={105086},
  year={2026},
  publisher={Elsevier}
}

@article{al2022open,
  title={Open-ended remote sensing visual question answering with transformers},
  author={Al Rahhal, Mohamad M and Bazi, Yakoub and Alsaleh, Sara O and Al-Razgan, Muna and Mekhalfi, Mohamed Lamine and Al Zuair, Mansour and Alajlan, Naif},
  journal={International Journal of Remote Sensing},
  volume={43},
  number={18},
  pages={6809--6823},
  year={2022},
  publisher={Taylor \& Francis}
}

@article{hu2025rsgpt,
  title={Rsgpt: A remote sensing vision language model and benchmark},
  author={Hu, Yuan and Yuan, Jianlong and Wen, Congcong and Lu, Xiaonan and Liu, Yu and Li, Xiang},
  journal={ISPRS Journal of Photogrammetry and Remote Sensing},
  volume={224},
  pages={272--286},
  year={2025},
  publisher={Elsevier}
}

@article{li2024vrsbench,
  title={Vrsbench: A versatile vision-language benchmark dataset for remote sensing image understanding},
  author={Li, Xiang and Ding, Jian and Elhoseiny, Mohamed},
  journal={Advances in Neural Information Processing Systems},
  volume={37},
  pages={3229--3242},
  year={2024}
}

@inproceedings{zhou2025urbench,
  title={Urbench: A comprehensive benchmark for evaluating large multimodal models in multi-view urban scenarios},
  author={Zhou, Baichuan and Yang, Haote and Chen, Dairong and Ye, Junyan and Bai, Tianyi and Yu, Jinhua and Zhang, Songyang and Lin, Dahua and He, Conghui and Li, Weijia},
  booktitle={Proceedings of the AAAI Conference on Artificial Intelligence},
  volume={39},
  number={10},
  pages={10707--10715},
  year={2025}
}

@inproceedings{licityanchor,
  title={CityAnchor: City-scale 3D Visual Grounding with Multi-modality LLMs},
  author={Li, Jinpeng and Wang, Haiping and Liu, Yuan and Dou, Zhiyang and Ma, Yuexin and Yang, Sibei and Li, Yuan and Wang, Wenping and Dong, Zhen and Yang, Bisheng and others},
  booktitle={The Thirteenth International Conference on Learning Representations}
}

@inproceedings{chen2021scan2cap,
  title={Scan2cap: Context-aware dense captioning in rgb-d scans},
  author={Chen, Zhenyu and Gholami, Ali and Nie{\ss}ner, Matthias and Chang, Angel X},
  booktitle={Proceedings of the IEEE/CVF conference on computer vision and pattern recognition},
  pages={3193--3203},
  year={2021}
}

@inproceedings{azuma2022scanqa,
  title={Scanqa: 3d question answering for spatial scene understanding},
  author={Azuma, Daichi and Miyanishi, Taiki and Kurita, Shuhei and Kawanabe, Motoaki},
  booktitle={proceedings of the IEEE/CVF conference on computer vision and pattern recognition},
  pages={19129--19139},
  year={2022}
}

@article{ma2022sqa3d,
  title={Sqa3d: Situated question answering in 3d scenes},
  author={Ma, Xiaojian and Yong, Silong and Zheng, Zilong and Li, Qing and Liang, Yitao and Zhu, Song-Chun and Huang, Siyuan},
  journal={arXiv preprint arXiv:2210.07474},
  year={2022}
}

@article{linghu2024multi,
  title={Multi-modal situated reasoning in 3d scenes},
  author={Linghu, Xiongkun and Huang, Jiangyong and Niu, Xuesong and Ma, Xiaojian Shawn and Jia, Baoxiong and Huang, Siyuan},
  journal={Advances in Neural Information Processing Systems},
  volume={37},
  pages={140903--140936},
  year={2024}
}

@inproceedings{chen2020scanrefer,
  title={Scanrefer: 3d object localization in rgb-d scans using natural language},
  author={Chen, Dave Zhenyu and Chang, Angel X and Nie{\ss}ner, Matthias},
  booktitle={European conference on computer vision},
  pages={202--221},
  year={2020},
  organization={Springer}
}

@inproceedings{wang2024embodiedscan,
  title={Embodiedscan: A holistic multi-modal 3d perception suite towards embodied ai},
  author={Wang, Tai and Mao, Xiaohan and Zhu, Chenming and Xu, Runsen and Lyu, Ruiyuan and Li, Peisen and Chen, Xiao and Zhang, Wenwei and Chen, Kai and Xue, Tianfan and others},
  booktitle={Proceedings of the IEEE/CVF Conference on Computer Vision and Pattern Recognition},
  pages={19757--19767},
  year={2024}
}

@inproceedings{yang20253d,
  title={3d-grand: A million-scale dataset for 3d-llms with better grounding and less hallucination},
  author={Yang, Jianing and Chen, Xuweiyi and Madaan, Nikhil and Iyengar, Madhavan and Qian, Shengyi and Fouhey, David F and Chai, Joyce},
  booktitle={Proceedings of the Computer Vision and Pattern Recognition Conference},
  pages={29501--29512},
  year={2025}
}

@article{miyanishi2023cityrefer,
  title={CityRefer: geography-aware 3D visual grounding dataset on city-scale point cloud data},
  author={Miyanishi, Taiki and Kitamori, Fumiya and Kurita, Shuhei and Lee, Jungdae and Kawanabe, Motoaki and Inoue, Nakamasa},
  journal={arXiv preprint arXiv:2310.18773},
  year={2023}
}

@inproceedings{sun20243d,
  title={3d question answering for city scene understanding},
  author={Sun, Penglei and Song, Yaoxian and Liu, Xiang and Yang, Xiaofei and Wang, Qiang and Li, Tiefeng and Yang, Yang and Chu, Xiaowen},
  booktitle={Proceedings of the 32nd ACM International Conference on Multimedia},
  pages={2156--2165},
  year={2024}
}

@inproceedings{qian2024nuscenes,
  title={Nuscenes-qa: A multi-modal visual question answering benchmark for autonomous driving scenario},
  author={Qian, Tianwen and Chen, Jingjing and Zhuo, Linhai and Jiao, Yang and Jiang, Yu-Gang},
  booktitle={Proceedings of the AAAI Conference on Artificial Intelligence},
  volume={38},
  number={5},
  pages={4542--4550},
  year={2024}
}

@inproceedings{xia2024text2loc,
  title={Text2loc: 3d point cloud localization from natural language},
  author={Xia, Yan and Shi, Letian and Ding, Zifeng and Henriques, Joao F and Cremers, Daniel},
  booktitle={Proceedings of the IEEE/CVF conference on computer vision and pattern recognition},
  pages={14958--14967},
  year={2024}
}

@article{feng2024citybench,
  title={Citybench: Evaluating the capabilities of large language model as world model},
  author={Feng, Jie and Zhang, Jun and Yan, Junbo and Zhang, Xin and Ouyang, Tianjian and Liu, Tianhui and Du, Yuwei and Guo, Siqi and Li, Yong},
  journal={arXiv e-prints},
  pages={arXiv--2406},
  year={2024}
}

@article{yasuki2025geoprog3d,
  title={GeoProg3D: Compositional Visual Reasoning for City-Scale 3D Language Fields},
  author={Yasuki, Shunsuke and Miyanishi, Taiki and Inoue, Nakamasa and Kurita, Shuhei and Sakamoto, Koya and Azuma, Daichi and Taki, Masato and Matsuo, Yutaka},
  journal={arXiv preprint arXiv:2506.23352},
  year={2025}
}

@article{sun2025city,
  title={City-VLM: Towards Multidomain Perception Scene Understanding via Multimodal Incomplete Learning},
  author={Sun, Penglei and Song, Yaoxian and Zhu, Xiangru and Liu, Xiang and Wang, Qiang and Liu, Yue and Xia, Changqun and Li, Tiefeng and Yang, Yang and Chu, Xiaowen},
  journal={arXiv preprint arXiv:2507.12795},
  year={2025}
}

@article{achiam2023gpt,
  title={Gpt-4 technical report},
  author={Achiam, Josh and Adler, Steven and Agarwal, Sandhini and Ahmad, Lama and Akkaya, Ilge and Aleman, Florencia Leoni and Almeida, Diogo and Altenschmidt, Janko and Altman, Sam and Anadkat, Shyamal and others},
  journal={arXiv preprint arXiv:2303.08774},
  year={2023}
}

@article{touvron2023llama,
  title={Llama: Open and efficient foundation language models},
  author={Touvron, Hugo and Lavril, Thibaut and Izacard, Gautier and Martinet, Xavier and Lachaux, Marie-Anne and Lacroix, Timoth{\'e}e and Rozi{\`e}re, Baptiste and Goyal, Naman and Hambro, Eric and Azhar, Faisal and others},
  journal={arXiv preprint arXiv:2302.13971},
  year={2023}
}

@article{ma2024llms,
  title={When llms step into the 3d world: A survey and meta-analysis of 3d tasks via multi-modal large language models},
  author={Ma, Xianzheng and Bhalgat, Yash and Smart, Brandon and Chen, Shuai and Li, Xinghui and Ding, Jian and Gu, Jindong and Chen, Dave Zhenyu and Peng, Songyou and Bian, Jia-Wang and others},
  journal={arXiv preprint arXiv:2405.10255},
  year={2024}
}

@inproceedings{xu2024pointllm,
  title={Pointllm: Empowering large language models to understand point clouds},
  author={Xu, Runsen and Wang, Xiaolong and Wang, Tai and Chen, Yilun and Pang, Jiangmiao and Lin, Dahua},
  booktitle={European Conference on Computer Vision},
  pages={131--147},
  year={2024},
  organization={Springer}
}

@article{wang2023chat,
  title={Chat-3d: Data-efficiently tuning large language model for universal dialogue of 3d scenes},
  author={Wang, Zehan and Huang, Haifeng and Zhao, Yang and Zhang, Ziang and Zhao, Zhou},
  journal={arXiv preprint arXiv:2308.08769},
  year={2023}
}

@article{zhu2024llava,
  title={Llava-3d: A simple yet effective pathway to empowering lmms with 3d-awareness},
  author={Zhu, Chenming and Wang, Tai and Zhang, Wenwei and Pang, Jiangmiao and Liu, Xihui},
  journal={arXiv preprint arXiv:2409.18125},
  year={2024}
}

@article{zhou2023uni3d,
  title={Uni3d: Exploring unified 3d representation at scale},
  author={Zhou, Junsheng and Wang, Jinsheng and Ma, Baorui and Liu, Yu-Shen and Huang, Tiejun and Wang, Xinlong},
  journal={arXiv preprint arXiv:2310.06773},
  year={2023}
}

@article{liu2023visual,
  title={Visual instruction tuning},
  author={Liu, Haotian and Li, Chunyuan and Wu, Qingyang and Lee, Yong Jae},
  journal={Advances in neural information processing systems},
  volume={36},
  pages={34892--34916},
  year={2023}
}

@article{qi2025gpt4scene,
  title={Gpt4scene: Understand 3d scenes from videos with vision-language models},
  author={Qi, Zhangyang and Zhang, Zhixiong and Fang, Ye and Wang, Jiaqi and Zhao, Hengshuang},
  journal={arXiv preprint arXiv:2501.01428},
  year={2025}
}

@inproceedings{zheng2025video,
  title={Video-3d llm: Learning position-aware video representation for 3d scene understanding},
  author={Zheng, Duo and Huang, Shijia and Wang, Liwei},
  booktitle={Proceedings of the Computer Vision and Pattern Recognition Conference},
  pages={8995--9006},
  year={2025}
}

@article{huang2024chat,
  title={Chat-scene: Bridging 3d scene and large language models with object identifiers},
  author={Huang, Haifeng and Chen, Yilun and Wang, Zehan and Huang, Rongjie and Xu, Runsen and Wang, Tai and Liu, Luping and Cheng, Xize and Zhao, Yang and Pang, Jiangmiao and others},
  journal={Advances in Neural Information Processing Systems},
  volume={37},
  pages={113991--114017},
  year={2024}
}

@article{hu2022lora,
  title={Lora: Low-rank adaptation of large language models.},
  author={Hu, Edward J and Shen, Yelong and Wallis, Phillip and Allen-Zhu, Zeyuan and Li, Yuanzhi and Wang, Shean and Wang, Lu and Chen, Weizhu and others},
  journal={ICLR},
  volume={1},
  number={2},
  pages={3},
  year={2022}
}

@article{liu2024deepseek,
  title={Deepseek-v3 technical report},
  author={Liu, Aixin and Feng, Bei and Xue, Bing and Wang, Bingxuan and Wu, Bochao and Lu, Chengda and Zhao, Chenggang and Deng, Chengqi and Zhang, Chenyu and Ruan, Chong and others},
  journal={arXiv preprint arXiv:2412.19437},
  year={2024}
}

@inproceedings{lai2024lisa,
  title={Lisa: Reasoning segmentation via large language model},
  author={Lai, Xin and Tian, Zhuotao and Chen, Yukang and Li, Yanwei and Yuan, Yuhui and Liu, Shu and Jia, Jiaya},
  booktitle={Proceedings of the IEEE/CVF Conference on Computer Vision and Pattern Recognition},
  pages={9579--9589},
  year={2024}
}

@article{thawakar2025llamav,
  title={Llamav-o1: Rethinking step-by-step visual reasoning in llms},
  author={Thawakar, Omkar and Dissanayake, Dinura and More, Ketan and Thawkar, Ritesh and Heakl, Ahmed and Ahsan, Noor and Li, Yuhao and Zumri, Mohammed and Lahoud, Jean and Anwer, Rao Muhammad and others},
  journal={arXiv preprint arXiv:2501.06186},
  year={2025}
}

@inproceedings{dong2025insight,
  title={Insight-v: Exploring long-chain visual reasoning with multimodal large language models},
  author={Dong, Yuhao and Liu, Zuyan and Sun, Hai-Long and Yang, Jingkang and Hu, Winston and Rao, Yongming and Liu, Ziwei},
  booktitle={Proceedings of the Computer Vision and Pattern Recognition Conference},
  pages={9062--9072},
  year={2025}
}

@inproceedings{radford2021learning,
  title={Learning transferable visual models from natural language supervision},
  author={Radford, Alec and Kim, Jong Wook and Hallacy, Chris and Ramesh, Aditya and Goh, Gabriel and Agarwal, Sandhini and Sastry, Girish and Askell, Amanda and Mishkin, Pamela and Clark, Jack and others},
  booktitle={International conference on machine learning},
  pages={8748--8763},
  year={2021},
  organization={PmLR}
}

@article{zhang2025open3dvqa,
  title={Open3dvqa: A benchmark for comprehensive spatial reasoning with multimodal large language model in open space},
  author={Zhang, Weichen and Zhou, Zile and Zheng, Zhiheng and Gao, Chen and Cui, Jinqiang and Li, Yong and Chen, Xinlei and Zhang, Xiao-Ping},
  journal={arXiv preprint arXiv:2503.11094},
  year={2025}
}

@article{zhao2025urbanvideo,
  title={Urbanvideo-bench: Benchmarking vision-language models on embodied intelligence with video data in urban spaces},
  author={Zhao, Baining and Fang, Jianjie and Dai, Zichao and Wang, Ziyou and Zha, Jirong and Zhang, Weichen and Gao, Chen and Wang, Yue and Cui, Jinqiang and Chen, Xinlei and others},
  journal={arXiv preprint arXiv:2503.06157},
  year={2025}
}

@article{yang2025qwen3,
  title={Qwen3 technical report},
  author={Yang, An and Li, Anfeng and Yang, Baosong and Zhang, Beichen and Hui, Binyuan and Zheng, Bo and Yu, Bowen and Gao, Chang and Huang, Chengen and Lv, Chenxu and others},
  journal={arXiv preprint arXiv:2505.09388},
  year={2025}
}

@article{liu2025logical,
  title={Logical reasoning in large language models: A survey},
  author={Liu, Hanmeng and Fu, Zhizhang and Ding, Mengru and Ning, Ruoxi and Zhang, Chaoli and Liu, Xiaozhang and Zhang, Yue},
  journal={arXiv preprint arXiv:2502.09100},
  year={2025}
}

@article{wang2024genartist,
  title={Genartist: Multimodal llm as an agent for unified image generation and editing},
  author={Wang, Zhenyu and Li, Aoxue and Li, Zhenguo and Liu, Xihui},
  journal={Advances in Neural Information Processing Systems},
  volume={37},
  pages={128374--128395},
  year={2024}
}

@inproceedings{ge2024worldgpt,
  title={Worldgpt: Empowering llm as multimodal world model},
  author={Ge, Zhiqi and Huang, Hongzhe and Zhou, Mingze and Li, Juncheng and Wang, Guoming and Tang, Siliang and Zhuang, Yueting},
  booktitle={Proceedings of the 32nd ACM International Conference on Multimedia},
  pages={7346--7355},
  year={2024}
}

@inproceedings{liu2024llava,
  title={Llava-plus: Learning to use tools for creating multimodal agents},
  author={Liu, Shilong and Cheng, Hao and Liu, Haotian and Zhang, Hao and Li, Feng and Ren, Tianhe and Zou, Xueyan and Yang, Jianwei and Su, Hang and Zhu, Jun and others},
  booktitle={European conference on computer vision},
  pages={126--142},
  year={2024},
  organization={Springer}
}

@article{awais2025foundation,
  title={Foundation models defining a new era in vision: a survey and outlook},
  author={Awais, Muhammad and Naseer, Muzammal and Khan, Salman and Anwer, Rao Muhammad and Cholakkal, Hisham and Shah, Mubarak and Yang, Ming-Hsuan and Khan, Fahad Shahbaz},
  journal={IEEE Transactions on Pattern Analysis and Machine Intelligence},
  year={2025},
  publisher={IEEE}
}

@article{xu2025pointllm,
  title={Pointllm-v2: Empowering large language models to better understand point clouds},
  author={Xu, Runsen and Yang, Shuai and Wang, Xiaolong and Wang, Tai and Chen, Yilun and Pang, Jiangmiao and Lin, Dahua},
  journal={IEEE Transactions on Pattern Analysis and Machine Intelligence},
  year={2025},
  publisher={IEEE}
}

@inproceedings{vu2022softgroup,
  title={Softgroup for 3d instance segmentation on point clouds},
  author={Vu, Thang and Kim, Kookhoi and Luu, Tung M and Nguyen, Thanh and Yoo, Chang D},
  booktitle={Proceedings of the IEEE/CVF conference on computer vision and pattern recognition},
  pages={2708--2717},
  year={2022}
}

@inproceedings{devlin2019bert,
  title={Bert: Pre-training of deep bidirectional transformers for language understanding},
  author={Devlin, Jacob and Chang, Ming-Wei and Lee, Kenton and Toutanova, Kristina},
  booktitle={Proceedings of the 2019 conference of the North American chapter of the association for computational linguistics: human language technologies, volume 1 (long and short papers)},
  pages={4171--4186},
  year={2019}
}

@article{comanici2025gemini,
  title={Gemini 2.5: Pushing the frontier with advanced reasoning, multimodality, long context, and next generation agentic capabilities},
  author={Comanici, Gheorghe and Bieber, Eric and Schaekermann, Mike and Pasupat, Ice and Sachdeva, Noveen and Dhillon, Inderjit and Blistein, Marcel and Ram, Ori and Zhang, Dan and Rosen, Evan and others},
  journal={arXiv preprint arXiv:2507.06261},
  year={2025}
}

@article{guo2025deepseek,
  title={Deepseek-r1: Incentivizing reasoning capability in llms via reinforcement learning},
  author={Guo, Daya and Yang, Dejian and Zhang, Haowei and Song, Junxiao and Zhang, Ruoyu and Xu, Runxin and Zhu, Qihao and Ma, Shirong and Wang, Peiyi and Bi, Xiao and others},
  journal={arXiv preprint arXiv:2501.12948},
  year={2025}
}

@article{hu2022sensaturban,
  title={Sensaturban: Learning semantics from urban-scale photogrammetric point clouds},
  author={Hu, Qingyong and Yang, Bo and Khalid, Sheikh and Xiao, Wen and Trigoni, Niki and Markham, Andrew},
  journal={International Journal of Computer Vision},
  volume={130},
  number={2},
  pages={316--343},
  year={2022},
  publisher={Springer}
}

@inproceedings{yang2023urbanbis,
  title={UrbanBIS: a large-scale benchmark for fine-grained urban building instance segmentation},
  author={Yang, Guoqing and Xue, Fuyou and Zhang, Qi and Xie, Ke and Fu, Chi-Wing and Huang, Hui},
  booktitle={ACM SIGGRAPH 2023 Conference Proceedings},
  pages={1--11},
  year={2023}
}

@inproceedings{zhu20233d,
  title={3d-vista: Pre-trained transformer for 3d vision and text alignment},
  author={Zhu, Ziyu and Ma, Xiaojian and Chen, Yixin and Deng, Zhidong and Huang, Siyuan and Li, Qing},
  booktitle={Proceedings of the IEEE/CVF International Conference on Computer Vision},
  pages={2911--2921},
  year={2023}
}

@article{zhao2023survey,
  title={A survey of large language models},
  author={Zhao, Wayne Xin and Zhou, Kun and Li, Junyi and Tang, Tianyi and Wang, Xiaolei and Hou, Yupeng and Min, Yingqian and Zhang, Beichen and Zhang, Junjie and Dong, Zican and others},
  journal={arXiv preprint arXiv:2303.18223},
  volume={1},
  number={2},
  year={2023}
}


\end{document}